\PassOptionsToPackage{table}{xcolor}
\documentclass[natbib]{pretty-preprint}

\newcommand{\FDCE}{\mathrm{FDCE}}

\usepackage{amsmath,amsfonts,bm}

\def\eqref#1{equation~\ref{#1}}

\def\1{\bm{1}}

\DeclareMathAlphabet{\mathsfit}{\encodingdefault}{\sfdefault}{m}{sl}
\SetMathAlphabet{\mathsfit}{bold}{\encodingdefault}{\sfdefault}{bx}{n}

\ifXeTeX
\fi

\usepackage{amssymb}
\usepackage{nicefrac}
\usepackage{microtype}
\ifPDFTeX\microtypesetup{expansion=true}\fi
\usepackage{xcolor}
\usepackage{graphicx}
\usepackage{booktabs}
\usepackage{multirow}
\usepackage{adjustbox}

\usepackage{array}
\usepackage{subcaption}
\usepackage{wrapfig}
\usepackage{float}
\usepackage{placeins}
\usepackage{afterpage}
\usepackage{hyperref}
\usepackage{url}

\newcommand{\eg}{e.g.,\xspace}
\usepackage{xspace}
\newcommand{\ie}{i.e.,\xspace}

\newcommand{\method}{CD-LAM}
\newcommand{\latstage}{ACWM debiased fine-tuning}
\newcommand{\Latstage}{ACWM Debiased Fine-tuning}
\newcommand{\ratstage}{robot action adaptation}
\newcommand{\Ratstage}{Robot Action Adaptation}
\newcommand{\fdce}{\textsc{FDCE}}

\title{Causally Debiased Latent Action Model for Embodied Action-Conditioned World Models}

\runningtitle{CD-LAM: Causally Debiased Latent Actions}
\author[1,2,\textdaggerdbl]{Yufan Wei}
\author[1,\textdagger]{Kun Zhou}
\author[1,2,\textdaggerdbl]{Lingjun Mao}
\author[1]{Ziming Xu}
\author[1,2,\textdaggerdbl]{Shuang Liang}
\author[1]{Zijun Zhang}
\author[1]{Ziqiao Xi}
\author[1]{Ruobing Han}
\author[1]{Yuchen Yan}
\author[1,2,\textdaggerdbl]{Xinyue Wang}
\author[1]{Fan Feng}
\author[1]{Biwei Huang}
\affiliation[1]{Aether AI}
\affiliation[2]{University of California, San Diego}
\contribution[\textdagger]{Project lead and corresponding author}
\contribution[\textdaggerdbl]{Done during an internship at Aether AI}
\correspondence{\email{franciskunzhou@gmail.com}}
\metadata[Code][\faGithub]{\href{https://github.com/AetherLabsAI/CD-LAM}{https://github.com/AetherLabsAI/CD-LAM}}
\metadata[Models][\huggingfaceicon]{\href{https://huggingface.co/AetherLabs-AI/CD-LAM}{https://huggingface.co/AetherLabs-AI/CD-LAM}}
\metadata[Project][\faGlobe]{\href{https://yufanwei.github.io/CD-LAM-project-page/}{yufanwei.github.io/CD-LAM-project-page}}
\metadata[Paper][\faFile]{\href{https://arxiv.org/abs/2607.09185}{arXiv:2607.09185}}

\renewcommand{\arraystretch}{1.15}

\renewcommand{\floatpagefraction}{0.7}
\abstract{

Action-conditioned world models (ACWMs) aim to simulate future observations conditioned on embodied actions. 
However, learning controllable ACWMs requires large-scale action-labeled data, which remains costly to collect. Latent action models (LAMs) mitigate this bottleneck by inferring latent actions from videos without executable action labels, but existing LAMs are typically trained with reconstruction-only objectives and therefore entangle action-relevant dynamics with action-irrelevant visual confounders. 
With such objectives, LAM favors visual context over action dynamics, therefore downstream ACWMs exhibit residual motion under zero actions and fail to reproduce supplied dynamics.
We propose CD-LAM, a causally debiased framework for LAM-based ACWMs. CD-LAM introduces three debiasing objectives: embodiment-centric reconstruction, action-centric contrastive learning, and latent space calibration, which together encourage embodiment-focused, action-aware, and well-calibrated representations. CD-LAM reduces action-following error by up to 42\% and 35\% when conditioned on latent and robot actions, while improving fidelity, and matching DreamDojo reference with over 12$\times$ fewer adaptation updates. CD-LAM also adapts LTX-2.3-22B into an ACWM with competitive quality using only 3.2\% of DreamDojo-14B's sample exposures. On X-VLA, pretraining with CD-LAM instead of baseline LAM consistently improves success rates by up to 5\% across LIBERO, LIBERO-Plus, and RoboTwin C2R.

}

\begin{document}

\maketitle

\section{Introduction}

Action-conditioned world models (ACWMs)~\citep{ha2018worldmodels,hafner2023dreamerv3,acwmphys2026,wu2024ivideogpt} have emerged as a promising paradigm for simulating the physical world directly from visual observations and embodied control signals. Given the current observation and an action trajectory, an ACWM aims to forecast the resulting future observation sequence, thereby simulating how the environment and embodiment would evolve under that intervention. By predicting futures under different candidate action trajectories, ACWMs have the potential to serve as general-purpose simulators for robot planning, policy evaluation, and data augmentation.
However, the controllability of ACWMs relies on large-scale action-labeled data, yet collecting robot videos with action annotations remains costly in the real world~\citep{khazatsky2024droid}. In contrast, human videos are easier to collect, far more scalable, and rich in physical interactions, but they typically lack executable action annotations~\citep{grauman2022ego4d,grauman2024egoexo4d,damen2018epic}. Latent action models (LAMs) offer a bridge to such large-scale data sources by inferring compact latent action representations from unlabeled video transitions~\citep{bruce2024genie,schmidt2024lapo,ye2024lapa}. Through large-scale pre-training conditioned on these latent actions, ACWMs can be effectively warmed up to learn action-controllable dynamics and achieve stronger action-following performance after fine-tuning on limited labeled robot data~\citep{gao2025adaworld,dreamdojo2026}.

Despite this promise, existing LAMs are typically trained solely with the video reconstruction objective, which cannot prevent action-irrelevant visual factors from being encoded into the latent actions. In particular, action-irrelevant factors such as background scenes and non-interacted objects may leak into the latent action representation. Our analysis in Section~\ref{sec:confounder} shows that such action-irrelevant factors induce biased latent action representations, which ultimately confound the ACWM: the model can generate visually plausible rollouts, but fails to reliably follow the supplied actions ~\citep{zhang2025whatdolams,nikulin2025latentdistractors,conla2026}. The inclusion of these confounders in the action condition can leave the ACWM with a biased state for robot-action adaptation and impose an additional alignment burden. We introduce latent-space diagnostics and intervention-based evaluations to visualize and quantify such bias. Our diagnostic results reveal substantial responses to small camera shifts and shortcuts that favor visual context over action similarity. Evaluation results further expose weak action following under both latent-action and robot-action conditioning, including residual motion under zero-action inputs and failure to reproduce target dynamics.



\begin{figure}[!t]
\centering
\includegraphics[width=1 \linewidth]{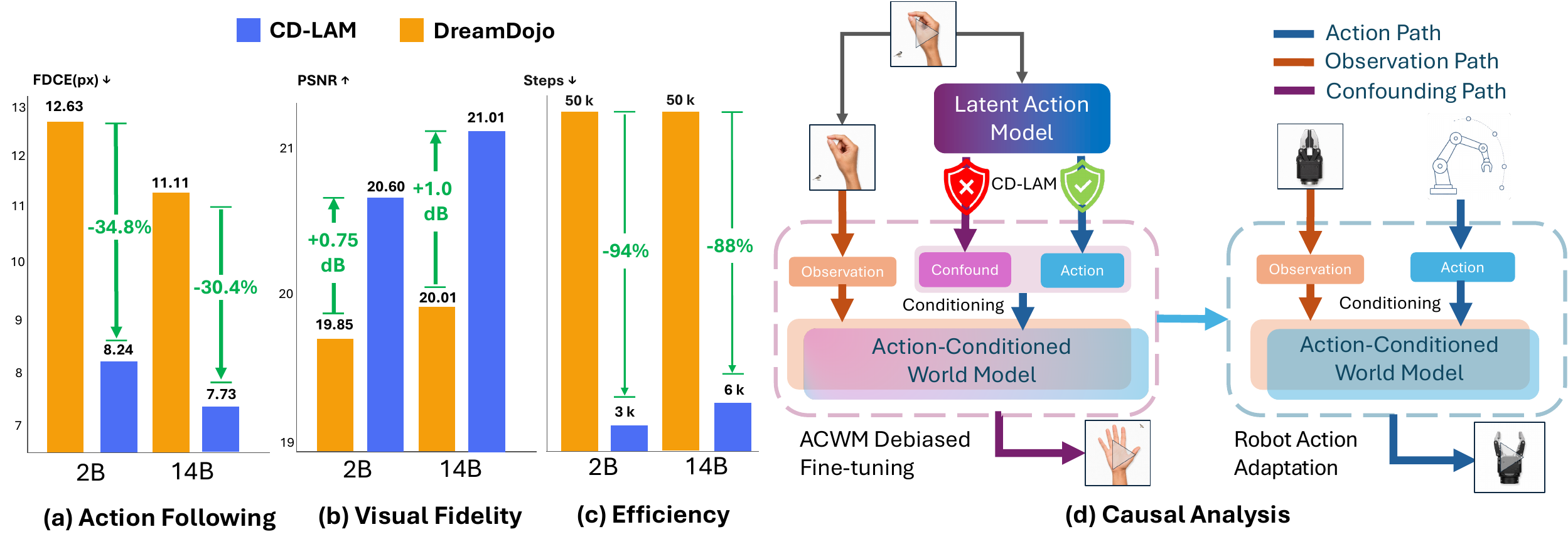}
\caption{\textbf{Performance overview and confounding analysis.}
CD-LAM lowers action-following error (FDCE) and raises PSNR at both 2B and 14B scales, with much lower training cost than DreamDojo. A reconstruction-trained LAM can leak action-irrelevant confounders into the action condition, and CD-LAM reduces this shortcut dependence.}
\label{fig:overall}\label{fig:causal}
\end{figure}%
Motivated by these findings, we propose CD-LAM to address action-irrelevant confounding in LAM-based ACWMs by explicitly shaping the latent action space. Specifically, it combines embodiment-centric reconstruction to emphasize the embodiment and interacted objects over background, action-centric contrastive learning to group transitions by coarse primitives across visual confounders and mitigate visual-context-driven shortcuts, and latent space calibration to suppress spurious responses to unchanged observations. CD-LAM consists of an efficient three-stage fine-tuning pipeline (Figure~\ref{fig:arch}) that first debiases the LAM, then the ACWM, and finally adapts the ACWM to real-world robot actions. In Stage 1, we fine-tune the LAM with these three debias objectives to emphasize embodiment-centric action dynamics. In Stage 2, we fine-tune the ACWM on debiased latent actions extracted from human videos without executable action labels, to debias the learned controllability and improve action following under the corrected latent conditions. In Stage 3, we add a lightweight action-to-latent-action mapping layer to bridge robot actions into the debiased latent space as the conditioning input and adapt the ACWM to executable robot controls. 



Experiments on 2B and 14B backbones show that upstream latent-action debiasing improves downstream world-model controllability (Figure~\ref{fig:overall}; Section~\ref{sec:experiments}). Before robot-action alignment, CD-LAM reduces action-following error, measured by Foreground Displacement Chamfer Error (FDCE), by 42\% and 26\%, respectively. These gains persist after \ratstage{}, with FDCE reductions of 35\% and 30\% alongside improved visual fidelity. The complete pipeline also matches the released DreamDojo reference with more than 12\(\times\) fewer \ratstage{} updates. Beyond DreamDojo, CD-LAM enables efficient adaptation of the pretrained 22B LTX-2.3 video backbone~\citep{lightricks2026ltx23} into an ACWM conditioned on executable robot actions. The resulting model achieves comparable rollout quality while using 3.2\% of the sample exposures reported for DreamDojo-14B across latent-action finetuning and robot-action adaptation. For policy learning, pretraining X-VLA~\citep{zheng2025xvla} with CD-LAM rather than the original LAM improves success rates across LIBERO, LIBERO-Plus and RoboTwin C2R~\citep{liu2023libero, fei2025liberoplus,chen2025robotwin}. 

This paper makes three contributions: 

$\bullet$ We identify and analyze a confounding mechanism in LAM-based ACWMs: action-irrelevant visual factors enter the latent action condition and undermine downstream controllability. 

$\bullet$ We propose CD-LAM, a causally motivated framework that targets this bias at its source by learning debiased action representations for world-model fine-tuning and executable robot-action alignment.

$\bullet$ 
Our CD-LAM yields persistent improvements on action-conditioned world models DreamDojo-2B/14B and LTX-2.3-22B, also improving policy models in three robot manipulation benchmarks.

\section{Preliminaries}
\label{sec:prelim}

\noindent\textbf{Action-conditioned world models.}\label{sec:prelim-acwm} An action-conditioned world model~(ACWM) predicts future observations under an action sequence:
 \begin{equation}
    p_\theta(o_{t+1:t+H}\mid o_{\leq t}, u_{t:t+H-1}),
    \label{eq:acwm-real-action}
\end{equation}
where \(o_t\) denotes the visual observation at time \(t\) and \(u_t\) denotes a recorded executable robot action, such as a relative end-effector displacement.

\noindent\textbf{Latent action models.}\label{sec:prelim-lam}
Given adjacent frames $\langle o_t,o_{t+1} \rangle$, the LAM infers a latent action vector \(z_t \in \mathbb{R}^d\), \ie a summary of the observed action-related information~\citep{bruce2024genie,schmidt2024lapo,ye2024lapa}.
A LAM can be written as an encoder paired with a decoder:
\begin{equation}
    z_t \sim q_\phi(z\mid o_t,o_{t+1}),
    \qquad
    \hat{o}_{t+1}=D_\psi(o_t,z_t).
    \label{eq:lam-autoencoder}
\end{equation}
We write \(\mu_\phi(o_t,o_{t+1})\) for the posterior mean of \(q_\phi\), used whenever a deterministic latent action is needed.
The standard LAM objective is next-frame reconstruction, optionally with a bottleneck or prior regularizer:
\begin{equation}
    \mathcal{L}_{\mathrm{LAM}}
    =
    \mathbb{E}\big[
        \ell(D_\psi(o_t,z_t),o_{t+1})
    \big]
    +
    \mathrm{Reg}(z_t).
    \label{eq:lam-loss}
\end{equation}
Here \(\ell\) is the observation reconstruction loss, and \(\mathrm{Reg}\) denotes the capacity-control mechanism used by the particular LAM, such as a KL bottleneck. This objective defines $z_t$ by information that helps reconstruct or predict the observed transition.

\noindent\textbf{Latent action conditioning.}\label{sec:prelim-latent-acwm} Prior LAM-based ACWM systems use latent actions as substitutes for missing robot actions during ACWM training on egocentric video: transitions are encoded as \((o_t,o_{t+1}) \mapsto z_t\), and the world model is trained as
\begin{equation}
    p_\theta(o_{t+1:t+H}\mid o_{\leq t}, z_{t:t+H-1}).
    \label{eq:latent-acwm}
\end{equation}
This pipeline requires two properties. The latent action should capture the embodied action in the observed transition, and the world model should make future motion follow the latent action condition~\citep{gao2025adaworld,dreamdojo2026}.  
Note that \(z_t\) plays a different role from \(u_t\): a recorded robot action \(u_t\) is an executable control signal defined by an embodiment and its controller, whereas \(z_t\) is a video-derived conditioning variable defined solely by the LAM training objective.

\section{Confounder Analysis on Latent Actions}
\label{sec:confounder}
In a LAM-based ACWM, \(z_t\) is the only action signal the world model receives during ACWM training, so any bias in \(z_t\) becomes a bias of the action condition itself. We first analyze why the reconstruction objective admits such bias, then verify the resulting failures in existing ACWMs.

\subsection{Why \texorpdfstring{a }{a }Reconstruction-centric Loss Admits Confounders}
For embodied ACWMs, \(z_t\) should act as the control signal whose dominant variation reflects key embodiment dynamics: body and object motion, contact, and displacement. A reconstruction-trained LAM does not enforce this property. Its objective only requires predictive sufficiency, \ie{} that \(z_t\) helps model \(p(o_{t+1}\mid o_t,z_t)\), so any transition-predictive factor can enter the latent:
\begin{equation}
    z_t = \mu_\phi(o_t,o_{t+1})
    \approx f(A_t, C_t, V_t),
    \label{eq:confounded-action-latent}
\end{equation}
where \(A_t\) denotes embodied action effects, \(C_t\) denotes scene context, and \(V_t\) denotes source-side visual factors such as background continuation and appearance continuity. The notation \(f(\cdot)\) records that nothing in the objective prevents \(z_t\) from carrying non-trivial dependence on \(C_t\) and \(V_t\) in addition to \(A_t\).
This induces \textbf{\textit{action-irrelevant confounding}}, illustrated in Figure~\ref{fig:causal}(d). Scene context and source-side visual factors are valid evidence in the observation history, where they help render the current scene. Once encoded into \(z_t\), however, they enter the action side of the model, and the condition partly specifies video continuation rather than embodiment dynamics. Robot action adaptation then inherits this contaminated latent space. A robot action \(u_t\) can ground the embodied dynamics in \(A_t\), but it cannot specify source-specific context, background continuation, or camera-like variation, so executable robot actions are forced into alignment with a partly non-actionable latent. This mismatch can increase the difficulty of robot-action alignment.

\subsection{Empirical Study on a Representative LAM-based ACWM}
\label{sec:interface-definitions}
\label{sec:observation}

We next check whether the analyzed failure appears in practice, using DreamDojo~\citep{dreamdojo2026} as a representative LAM-based ACWM. The evidence has two layers:

\noindent\textbf{{Weak Robot Action Following.}} After \ratstage{}, the generated embodiment dynamics should follow the conditioning action. However, DreamDojo ACWMs tend to violate this requirement in two basic settings, both of which keep the initial frame fixed and replace only the conditioning action. 
First, in the zero-action setting, we replace the conditioning action with zero, denoted \(\mathrm{do}(u_t=0)\). Although the embodiment motion should be suppressed, the rollout keeps moving (Figure~\ref{fig:zeroframe}). Second, under target-action transfer, where the conditioning action is replaced with one taken from a different target video, the rollout should reproduce the target embodiment dynamics, yet it does not (Figure~\ref{fig:transferaction}).
These examples indicate that the ACWM does not reliably follow the supplied robot actions. Both settings are interventions in the causal sense: the observation history is held fixed and only the action condition is replaced. These probes test the model's response to replacing the action condition while keeping the observation fixed, turning the disentanglement that Equation~(\ref{eq:confounded-action-latent}) calls for into a behavioral test, and Section~\ref{sec:exp-posttrain} applies the same two interventions.

\noindent\textbf{{Action-irrelevant Confounding in Latent Actions.}} We further examine the latent action space directly. A clean latent action space should satisfy three basic properties: (1) the zero-transition latent $z_t^{0}=\mu_\phi(o_t,o_t)$, encoded from a duplicated-frame pair, should remain close to zero; (2) camera-like image shifts should induce only small latent responses; and (3) local neighborhoods should reflect action similarity rather than shared visual context.  To quantify these three properties, we design corresponding evaluation metrics  (Appendix~\ref{app:metrics}) and apply them to the DreamDojo LAM encoder. As shown in Table~\ref{tab:latent-audit}, the DreamDojo LAM falls short of these desired properties and exhibits substantial bias: duplicated-frame pairs and synthetic camera shifts produce large latent responses, while same-episode visual context pulls latents closer even when their action primitives differ. The DreamDojo latent actions are thus significantly biased by action-irrelevant confounders.

\begin{figure}[!t]
\captionsetup{font=footnotesize}
\noindent\begin{minipage}[t]{0.49\linewidth}
\vspace{0pt}
\begin{subfigure}{\linewidth}
\centering
\includegraphics[width=\linewidth]{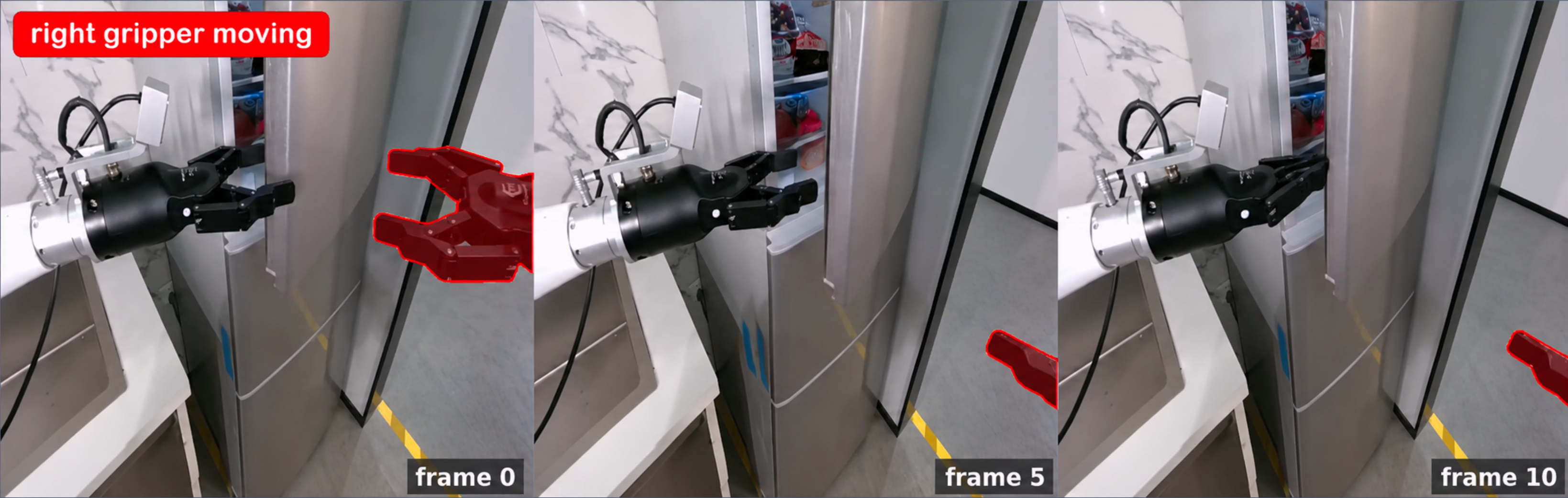}
\caption{Zero-action input.}
\label{fig:zeroframe}
\end{subfigure}

\vspace{8pt}
\captionsetup{type=table,position=above,justification=raggedright,singlelinecheck=false,skip=3pt}
\caption{\textbf{LAM encoder confounding audit}. A clean LAM encoder barely responds to action irrelevant changes (more details at  Table~\ref{tab:latent-audit-full}).}
\label{tab:latent-audit}
{\fontsize{8}{10}\selectfont
\setlength{\tabcolsep}{2pt}
\renewcommand{\arraystretch}{1.05}
\begin{tabular*}{\linewidth}{@{}l@{\extracolsep{\fill}}cc@{}}
\toprule
Diagnostic & DreamDojo & \method{} \\
\midrule
\multicolumn{3}{@{}l}{\emph{Zero-transition} (static pair $(o_t,o_t)$)}\\
\quad Rel.\ norm, median$\downarrow$ & $0.527$ & $\mathbf{0.043}$ \\
\quad Abs.\ norm, median$\downarrow$ & $3.119$ & $\mathbf{0.226}$ \\
\addlinespace[1pt]
\multicolumn{3}{@{}l}{\emph{Camera shift} (rel.\ norm, mean/median)}\\
\quad Horizontal$\downarrow$ & $0.555/0.536$ & $\mathbf{0.156/0.096}$ \\
\quad Vertical$\downarrow$ & $0.545/0.529$ & $\mathbf{0.110/0.064}$ \\
\addlinespace[1pt]
\multicolumn{3}{@{}l}{\emph{Action-neighbor structure}}\\
\quad Shortcut leakage$\downarrow$ & $0.151$ & $\mathbf{0.014}$ \\
\bottomrule
\end{tabular*}\par}
\end{minipage}\hfill
\begin{minipage}[t]{0.475\linewidth}
\setcounter{subfigure}{1}
\vspace{0pt}
\begin{subfigure}{\linewidth}
\centering
\includegraphics[width=\linewidth]{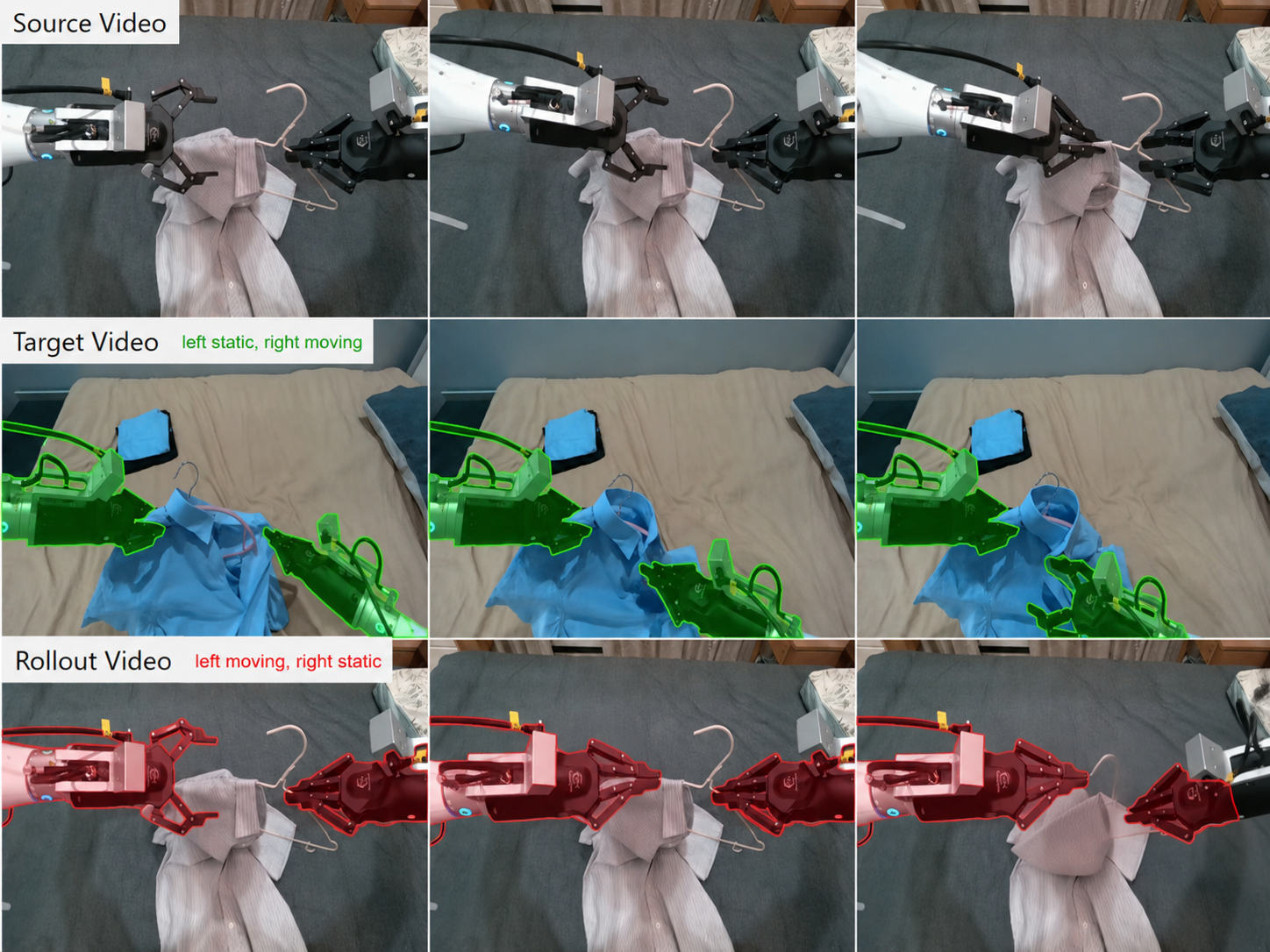}
\caption{Transferred target action.}
\label{fig:transferaction}
\end{subfigure}

\vspace{4pt}
\captionsetup{type=figure,position=below,justification=raggedright,singlelinecheck=false}
\addtocounter{figure}{-1}
\caption{\textbf{Action-following failures under zero-input and target-action probes.} The initial frame is fixed and only the action is replaced. (a)~Zero action: the rollout still moves. (b)~Actions transferred from a target video (middle, green): the rollout (bottom, red) misses the target dynamics.}
\label{fig:action-intervention-failures}
\end{minipage}
\end{figure}

\section{CD-LAM: Causal Debiasing of Latent Actions}
\label{sec:method}
Section~\ref{sec:confounder} shows that reconstruction-only LAM training can confound latent actions. We propose CD-LAM, a causally motivated debiasing approach with an efficient staged fine-tuning pipeline.
\begin{figure}[!hbtp]
\centering
\includegraphics[width=1\linewidth]{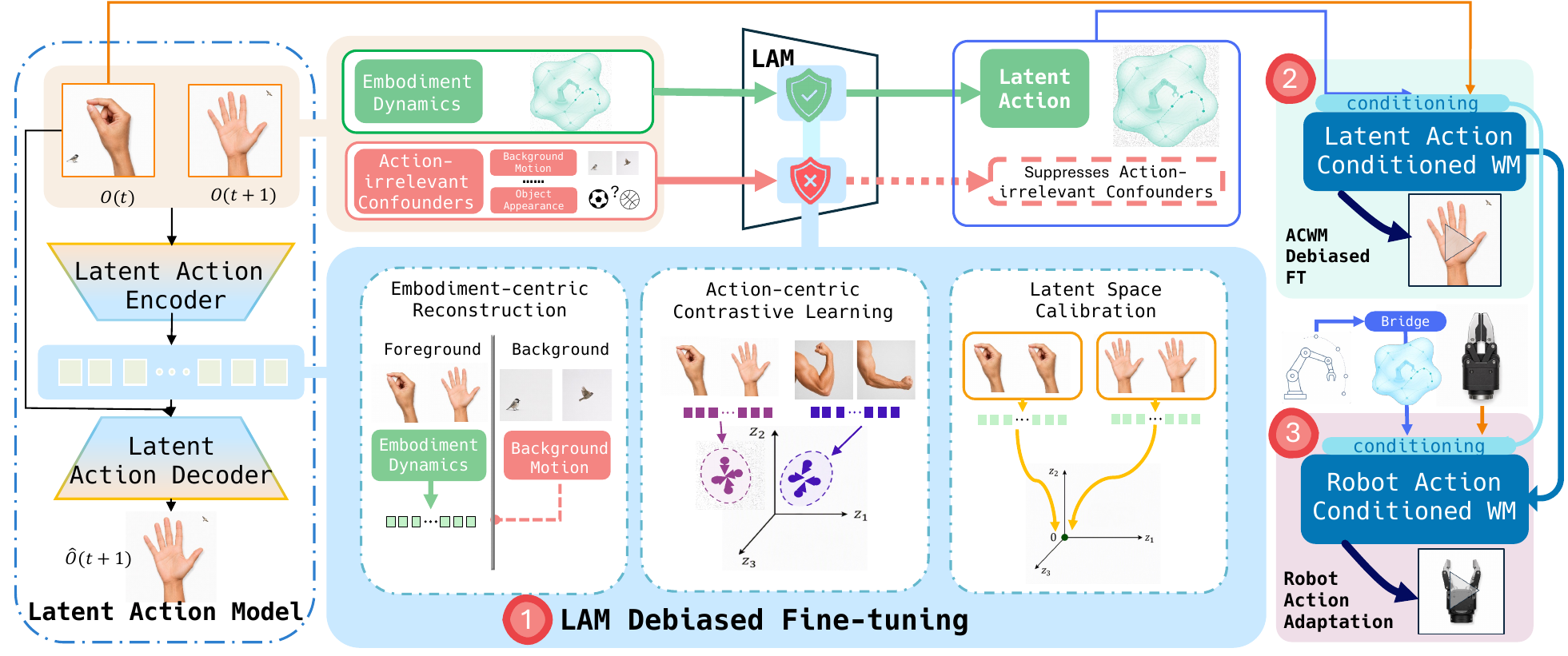}
\caption{\textbf{Overview of CD-LAM.}
CD-LAM debiases the LAM's latent action space in three stages while keeping the downstream action conditioning format unchanged. \textbf{Stage 1 (LAM debiased fine-tuning)} debiases the LAM with the three CD-LAM objectives. \textbf{Stage 2 (\latstage{})} trains the ACWM on the debiased latent actions. \textbf{Stage 3 (\ratstage{})} aligns executable robot actions to the same space through a lightweight bridge.}
\label{fig:arch}
\end{figure}

\subsection{Design Principles}\label{sec:method-principles}
The analysis in Section~\ref{sec:confounder} leads to three design principles for keeping action-irrelevant factors out of the latent action space while visual context stays available through the observation path.

$\bullet$ \textbf{Embodiment-centric Reconstruction.} The reconstruction signal should emphasize regions tied to embodiment dynamics over background regions. \\
$\bullet$ \textbf{Action-centric Structure.} The latent action space should group videos with similar actions, not visually similar videos with different actions. \\
$\bullet$ \textbf{Calibrated Latent Space.} Duplicated-frame inputs should map to a designated zero-transition reference, while ordinary transitions retain enough variation to encode embodiment dynamics. 

CD-LAM implements these principles with three matching LAM objectives: embodiment-centric reconstruction (\(\mathcal L_{\mathrm{emb}}\)), action-centric contrastive learning (\(\mathcal L_{\mathrm{ctr}}\)), and latent space calibration (\(\mathcal L_{\mathrm{cal}}\)).

\subsection{CD-LAM Objective}

Given a transition \((o_t,o_{t+1})\), CD-LAM produces a debiased latent action
\begin{equation}
z_t^{\mathrm{CD}}=\mu_\phi(o_t,o_{t+1})\in\mathbb{R}^{d},
\label{eq:zcd}
\end{equation}
which is fed into the same ACWM conditioning format as the original DreamDojo latent.
The LAM is fine-tuned with
\begin{equation}
    \mathcal L_{\mathrm{CD}}
    =
    \mathcal L_{\mathrm{emb}}
    + \lambda_{\mathrm{ctr}}(k)\mathcal L_{\mathrm{ctr}}
    + \mathcal L_{\mathrm{cal}} .
    \label{eq:cdlam-total}
\end{equation}
 Here \(\lambda_{\mathrm{ctr}}(k)\) weights the contrastive term and varies with the training step \(k\) (we reserve \(t\) for frame time); the calibration term carries its own weights (Equation~(\ref{eq:action-norm-compact})).

\subsubsection{Embodiment-centric Reconstruction Loss}

We use an embodiment-centric weighted MSE for frame-pair reconstruction. Given a transition \((o_t,o_{t+1})\), CD-LAM predicts \(\hat{o}_{t+1}=D_\psi(o_t,z_t^{\mathrm{CD}})\) as in Equation~(\ref{eq:lam-autoencoder}).
Let \(M_t\in[0,1]^{h\times w}\) be the embodiment--object foreground mask
obtained by SAM3~\citep{carion2025sam3}, where \(h\times w\) is the frame resolution. We optimize the weighted reconstruction loss
{\small
\begin{equation}
    \mathcal L_{\mathrm{emb}}
    = \frac{1}{|\Omega|}\sum_{p\in\Omega}
    W_t(p)\,\big\|\hat{o}_{t+1}(p)-o_{t+1}(p)\big\|_2^2,
    \qquad
    W_t = \alpha_{\mathrm{fg}} M_t
        + \alpha_{\mathrm{bg}}(1-M_t).
    \label{eq:emb-rec}
\end{equation}}
where \(\Omega\) is the pixel grid and \(\alpha_{\mathrm{fg}}>\alpha_{\mathrm{bg}}\): embodiment and object errors cost more than background errors, while the nonzero background weight keeps global visual consistency.

\subsubsection{Action-centric Contrastive Learning}
Reconstruction alone treats each transition independently. CD-LAM adds a pairwise contrastive loss over coarse manipulation primitives so that action-consistent transitions remain close across visual contexts. Let \(v_i=\mathrm{norm}(r_\omega(z_i^{\mathrm{CD}}))\) with an auxiliary head \(r_\omega\), let \(s_{ij}=\tau\,v_i^\top v_j+b\) be their scaled cosine similarity, and let \(\mathcal P_\pm\) be the valid same-/different-primitive pairs in a batch:
{\small
\begin{equation}
\mathcal L_{\mathrm{ctr}}
=
\underbrace{\frac{1}{W_+}\sum_{(i,j)\in\mathcal P_+} w_{ij}\,\mathrm{softplus}(-s_{ij})}_{\text{pull same primitive together}}
+
\underbrace{\frac{1}{W_-}\sum_{(i,j)\in\mathcal P_-} w_{ij}\,\mathrm{softplus}(s_{ij})}_{\text{push different primitives apart}} .
\label{eq:contrast-compact}
\end{equation}
}

Each term is normalized by its own total pair weight, \(W_\pm=\max\!\big(1,\sum_{(i,j)\in\mathcal P_\pm}w_{ij}\big)\), and batches without positives contribute zero. The learned scale \(\tau\), bias \(b\)~\citep{zhai2023siglip}, and pair weights \(w_{ij}\) are specified in the released configuration.
The primitive labels are coarse verb-level categories (\eg{} pick--place, pour, open, close) obtained by mapping the datasets' task and verb annotations into a coarse primitive space (Appendix~\ref{app:verbs}) without involving any executable robot actions. Transitions without annotations still contribute to the reconstruction and calibration objectives.

\subsubsection{Latent Space Calibration}
The latent space calibration loss combines capacity control with zero-transition calibration:

\begin{equation}
    \mathcal L_{\mathrm{cal}}
    =
    \beta_{\mathrm{kl}}(k)\,\mathcal L_{\mathrm{KL\text{-}fb}}
    +
    \lambda_{\mathrm{zero}}\,\mathcal L_{\mathrm{zero}} ,~\quad
    \mathcal L_{\mathrm{zero}}
    =
    \mathbb E_{o_t}\big[\,[\rho_t-m_{\mathrm{zero}}]_{+}^{2}\,\big] .
    \label{eq:action-norm-compact}
\end{equation}

\noindent
The free-bit KL term \(\mathcal L_{\mathrm{KL\text{-}fb}}\) is a KL penalty applied only above a per-dimension free-bits floor. It provides capacity control while preserving variation among ordinary transitions. The zero-transition term \(\mathcal L_{\mathrm{zero}}\) acts on the relative norm \mbox{\(\rho_t=\|z_t^{0}\|_2/(\mathrm{sg}(s_\Delta)+\epsilon)\)} of the zero-transition latent \(z_t^{0}=\mu_\phi(o_t,o_t)\) (Section~\ref{sec:observation}), where \(s_\Delta\) is the running RMS norm of ordinary transition latents. It is nonzero only when \(\rho_t\) exceeds a small margin \(m_{\mathrm{zero}}\), so duplicated-frame inputs are encouraged toward the designated zero-transition reference. Here \(\mathrm{sg}(\cdot)\) is stop-gradient, \([x]_+=\max(x,0)\), and \(\beta_{\mathrm{kl}}(k)\), \(\lambda_{\mathrm{zero}}\) are term weights (Table~\ref{tab:loss}).

\subsection{Multi-stage Training}\label{sec:method-bridge}

CD-LAM is used in a three-stage training pipeline (Figure~\ref{fig:arch}):

$\bullet$ \textbf{Stage 1 (LAM Debiased Fine-tuning).} We first fine-tune the LAM on videos without executable action labels using the CD-LAM objective of Equation~(\ref{eq:cdlam-total}) with only coarse task annotation. 

$\bullet$ \textbf{Stage 2 (ACWM Debiased Fine-tuning).} We then extract debiased latents \(z_t^{\mathrm{CD}}\) from videos and continue ACWM training under the conditioning of Equation~(\ref{eq:latent-acwm}), with \(z_t^{\mathrm{CD}}\) in place of \(z_t\). This adapts the world model to the debiased latent action space before executable robot actions are introduced.

$\bullet$ \textbf{Stage 3 (\Ratstage).}
Finally, for action labeled robot data \((o_t,u_t,o_{t+1})\), we align actions to the debiased latent action space. A lightweight MLP bridge \(g_\eta\) is trained to regress the gradient-stopped CD-LAM latent, \(g_\eta(u_t)\approx\mathrm{sg}\big(\mu_\phi(o_t,o_{t+1})\big)\). An auxiliary action readout with a cycle-consistency objective encourages the mapped latent to retain information about \(u_t\). During \ratstage{}, the ACWM receives \(\hat z_t=g_\eta(u_t)\), so robot actions enter the same debiased latent action space used in Stage 2. 

\section{Experiments}
\label{sec:experiments}\label{sec:exp}\label{sec:scale}

\subsection{Experimental Setup}
\label{sec:exp-axes}
\textbf{Models and baseline.}
We apply CD-LAM to DreamDojo~\citep{dreamdojo2026} at the 2B and 14B scales, which share a single latent action model, while keeping the downstream ACWM architecture, latent dimensionality, and conditioning interface unchanged. DreamDojo serves as the matched baseline to isolate the effect of latent-action debiasing within the same LAM--ACWM pipeline. For Stage 2, we continue the original DreamDojo training on the same video subset with matched update counts. For Stage 3, we compare against its released robot-action-adapted checkpoints under the same evaluation protocol. Recent latent-action methods span different levels of the learning stack, from alternative latent formulations and joint latent/world-model learning to policy/VLA pretraining~\citep{colaworld2025,ayalew2026claw,zhang2026dila}, so their reported scores also reflect differences in downstream architecture, action grounding, embodiment, and evaluation. We therefore use DreamDojo for the debiasing study, while LTX-2.3 and X-VLA serve as complementary transfer studies evaluating whether the action-conditioning interface and the learned latent representation remain useful beyond DreamDojo pipeline~\citep{hacohen2026ltx2,zheng2025xvla}.

\textbf{Training and data.} The main study runs on 128 H100 GPUs. Our three-stage fine-tuning starts from released DreamDojo checkpoints. By default, LAM debiasing uses 100h of DreamDojo pretraining human video, and robot-action adaptation uses approximately 300h of paired AgiBot video and actions~\citep{agibotworldcontributors2025agibotworldcolosseolargescale}. We evaluate latent-action rollouts on EgoDex~\citep{hoque2026egodexlearningdexterousmanipulation} and robot-action rollouts on held-out AgiBot clips. Training and evaluation details are provided in Appendix~\ref{app:setup}, with separate LTX and X-VLA protocols in Appendices~\ref{app:ltx} and~\ref{app:vla}.

\begin{figure}[!htbp]
\captionsetup{font=footnotesize,skip=5pt,justification=raggedright,singlelinecheck=false}
\noindent\begin{minipage}[t]{.49\linewidth}\vspace{0pt}
\captionsetup{justification=raggedright,singlelinecheck=false}
\centering

\includegraphics[width=0.70\linewidth]{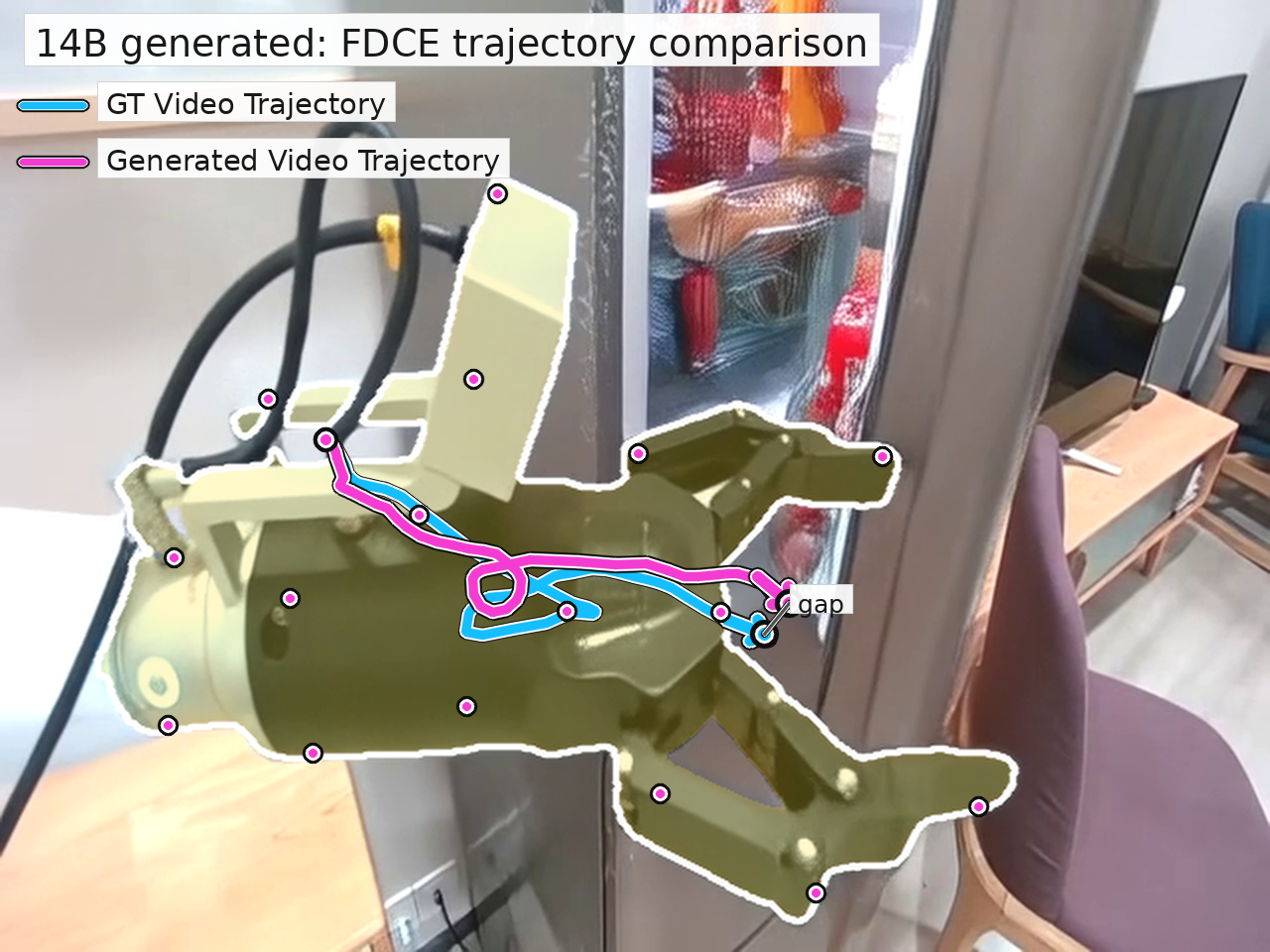}
\caption{\textbf{\fdce{} computation:} symmetric Chamfer distance of foreground displacement tracks.}
\label{fig:metric-explainer}
\end{minipage}\hfill\begin{minipage}[t]{.48\linewidth}\vspace{0pt}
\centering
\includegraphics[width=0.74\linewidth]{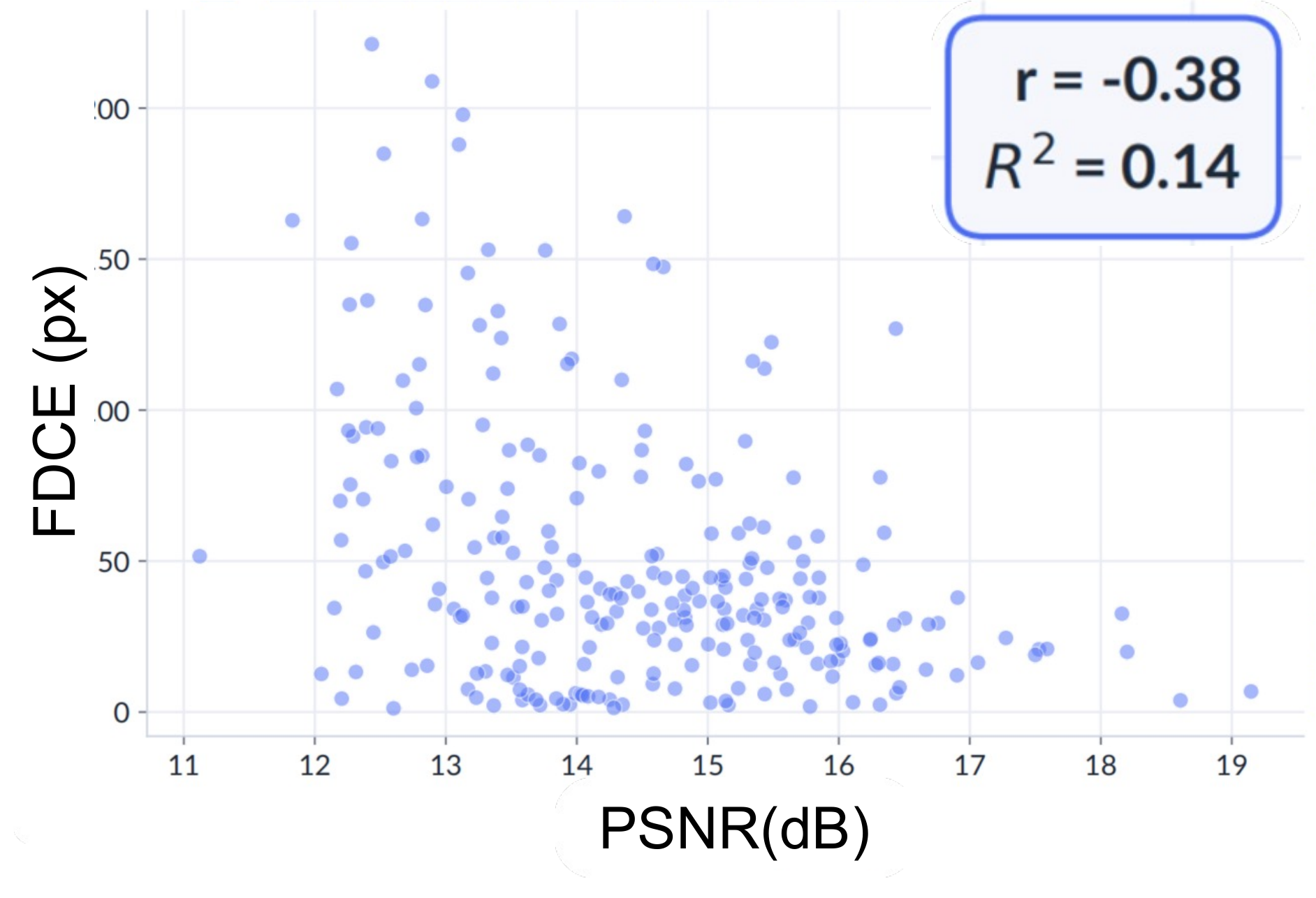}
\caption{\textbf{Pixel-level fidelity does not certify action following.} PSNR explains only limited variance in FDCE (\(r=-0.38\), \(R^2=0.14\)). Each point is one rollout.}
\label{fig:psnr-fdce}
\end{minipage}
\end{figure}

\textbf{Metrics.} We evaluate visual fidelity with PSNR, SSIM~\citep{wang2004ssim}, and LPIPS~\citep{zhang2018perceptual}, and action following with Foreground Displacement Chamfer Error (\fdce{}). Pixel metrics alone do not certify action following: as Figure~\ref{fig:psnr-fdce} shows, high PSNR can coexist with incorrect embodiment motion. Following trajectory-based motion evaluation for controllable video generation~\citep{motionctrl2024,motionpro2025}, \fdce{} measures the symmetric Chamfer distance between generated and reference foreground displacement tracks in pixels; Figure~\ref{fig:metric-explainer} illustrates the computation and Appendix~\ref{app:metrics} provides the full definition. As a tracking-based proxy for action following, we interpret \fdce{} jointly with zero-action and target-action interventions, visual-fidelity metrics, and qualitative rollouts. For VLA policies, we report success rate.

\subsection{Latent Action Diagnostics (Stage 1)}
\label{sec:exp-lam}\label{sec:audit}%
We first audit the latent action before any world model rollout. This directly probes the upstream source identified in Section~\ref{sec:confounder}: whether the encoder responds to action-irrelevant confounders as if they were embodied action. Table~\ref{tab:latent-audit} shows that \method{} strongly reduces responses to action-irrelevant confounders: the median zero-transition response drops from 0.527 to 0.043, synthetic camera shifts induce 3.6--8.3\(\times\) smaller responses. Shortcut leakage is computed as the difference between the mean cosine similarity of same-episode pairs with different primitives and that of different-episode pairs with the same primitive. It drops from 0.151 to 0.014, indicating substantially less context-driven structure in the latent space.

\begin{table}[!htbp]
\caption{\textbf{Latent action conditioned rollouts after \latstage.}
Left: conditioned on \(z_t\) from the video transition. Right: target-action transfer \(\mathrm{do}(z_t=z_t^{\mathrm{tar}})\).}
\label{tab:wmego}\label{tab:wm-egodex}

\centering\footnotesize
\setlength{\tabcolsep}{3.2pt}\renewcommand{\arraystretch}{1.1}
\begin{adjustbox}{max width=\linewidth}
\begin{tabular}{@{}llcccccccc@{}}
\toprule
& & \multicolumn{4}{c}{latent action \(z_t\) rollout} & \multicolumn{4}{c}{ \(\mathrm{do}(z_t=z_t^{\mathrm{tar}})\) } \\
\cmidrule(lr){3-6}\cmidrule(lr){7-10}
Backbone & Model & \fdce{}\,$\downarrow$ & PSNR\,$\uparrow$ & SSIM\,$\uparrow$ & LPIPS\,$\downarrow$
      & PSNR\,$\uparrow$ & SSIM\,$\uparrow$ & LPIPS\,$\downarrow$ & \fdce{}\,$\downarrow$ \\
\midrule
\multirow{2}{*}{2B}
 & DreamDojo      & $34.00$ & $20.88$ & $0.780$ & $0.413$ & $13.02$ & $\mathbf{0.598}$ & $0.643$ & $42.74$ \\
 & \method{}  & $\mathbf{19.63}$ & $\mathbf{24.29}$ & $\mathbf{0.827}$ & $\mathbf{0.308}$ & $\mathbf{13.15}$ & $0.588$ & $\mathbf{0.600}$ & $\mathbf{33.81}$ \\
\midrule
\multirow{2}{*}{14B}
 & DreamDojo & $40.29$ & $21.04$ & $0.792$ & $0.398$ & $\mathbf{13.11}$ & $0.593$ & $0.631$ & $50.27$ \\
 & \method{}  & $\mathbf{29.87}$ & $\mathbf{23.18}$ & $\mathbf{0.814}$ & $\mathbf{0.342}$ & $13.03$ & $\mathbf{0.597}$ & $\mathbf{0.617}$ & $\mathbf{33.22}$ \\
\bottomrule
\end{tabular}
\end{adjustbox}
\end{table}

\subsection{\Latstage{} Rollouts (Stage 2)}
\label{sec:exp-pretrain}
We then evaluate ACWM rollouts conditioned on debiased LAM latents before introducing the robot-action bridge. Table~\ref{tab:wm-egodex} shows gains at two scales. Compared with the matched continued-training baseline, CD-LAM reduces FDCE by 42\% at 2B (34.00 to 19.63) and by 26\% at 14B (40.29 to 29.87) while also improving PSNR, SSIM, and LPIPS. Under target-action transfer \(\mathrm{do}(z_t=z_t^{\mathrm{tar}})\), FDCE drops by 21\% and 34\%, respectively, while pixel metrics show smaller, mixed changes. These results show that CD-LAM improves action following before robot-action alignment. We also found that scaling the baseline from 2B to 14B does not improve action following, whereas CD-LAM reduces FDCE at both scales.

\begin{table}[!htb]
\centering\footnotesize
\setlength{\tabcolsep}{3.2pt}\renewcommand{\arraystretch}{1.2}
\caption{\textbf{Robot action rollouts and interventions on AgiBot (Stage 3).} Zero-action FDCE is residual motion w.r.t.\ the static initial frame. DreamDojo: released 50k-update checkpoints.}
\label{tab:posttrain-recon-2b14b}\label{tab:ego}\label{tab:posttrain-main}\label{tab:interventions}\label{tab:latent-interventions}

\begin{adjustbox}{max width=\linewidth}
\begin{tabular}{@{}llccccccc@{}}
\toprule
& & \multicolumn{5}{c}{robot action \(u_t\) rollout} & \(\mathrm{do}(u_t=0)\) & \(\mathrm{do}(u_t=u_t^{\mathrm{tar}})\) \\
\cmidrule(lr){3-7}\cmidrule(lr){8-9}
Backbone & Model & FDCE$_{\text{mean}}$\,$\downarrow$ & FDCE$_{\text{med}}$\,$\downarrow$ & PSNR\,$\uparrow$ & SSIM\,$\uparrow$ & LPIPS\,$\downarrow$
 & FDCE\,$\downarrow$ & FDCE\,$\downarrow$ \\
\midrule
\multirow{2}{*}{2B}
& DreamDojo & $12.63$ & $8.15$ & $19.85$ & $0.798$ & $0.271$ & $10.71$ & $24.36$ \\
& \method{} & $\mathbf{8.24}$ & $\mathbf{6.75}$ & $\mathbf{20.60}$ & $\mathbf{0.806}$ & $\mathbf{0.269}$ & $\mathbf{5.03}$ & $\mathbf{22.55}$ \\
\midrule
\multirow{2}{*}{14B}
 & DreamDojo & $11.11$ & $8.98$ & $20.01$ & $0.808$ & $0.263$ & $9.36$ & $24.82$ \\
& \method{} & $\mathbf{7.73}$ & $\mathbf{5.99}$ & $\mathbf{21.01}$ & $\mathbf{0.818}$ & $\mathbf{0.247}$ & $\mathbf{2.18}$ & $\mathbf{21.11}$ \\
\bottomrule
\end{tabular}
\end{adjustbox}
\end{table}

\subsection{\Ratstage{} Rollouts (Stage 3)}
\label{sec:exp-posttrain}\label{sec:main}\label{sec:repair}\label{sec:exp-wm}\label{sec:reuse}\label{sec:exp-interventions}\label{sec:interv}

\begin{figure}[!htbp]
\centering

\includegraphics[width=1\linewidth]{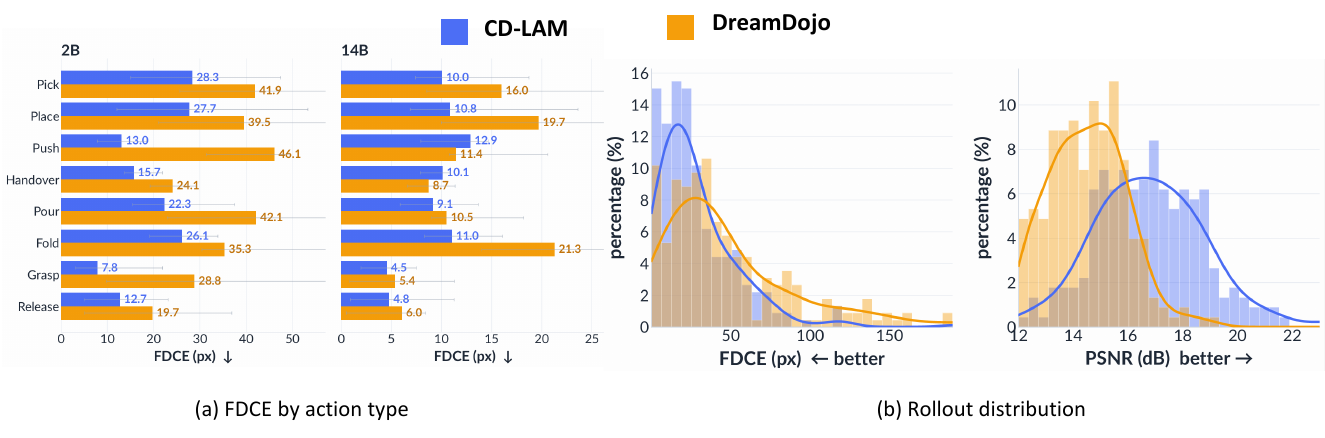}
\caption{\textbf{Action-following behavior after \ratstage{}, beyond aggregate scores.} (a) CD-LAM reduces FDCE in most action categories (full breakdown in Figure~\ref{fig:appendix-target-and-action}) at both 2B and 14B scales, so the gain is not concentrated in a single primitive. (b) On the 2B ACWM after \ratstage, CD-LAM shifts the rollout distribution toward lower FDCE and higher PSNR.}
\label{fig:posttrain-panels}
\end{figure}

\begin{figure}[!htb]
\noindent\begin{minipage}[c]{0.56\linewidth}
\centering
\includegraphics[width=\linewidth]{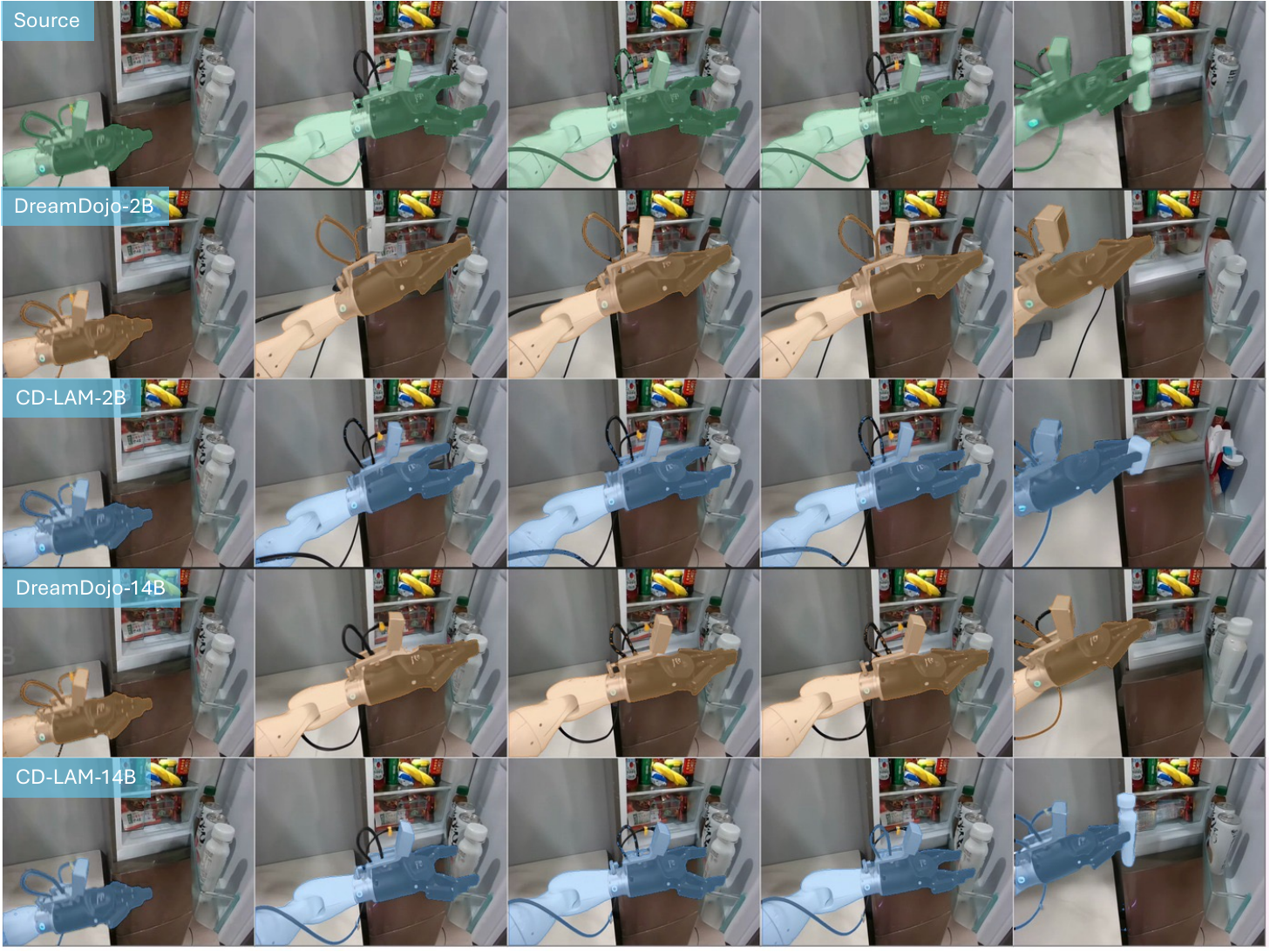}
\end{minipage}\hfill
\begin{minipage}[c]{0.41\linewidth}
\captionsetup{font=footnotesize,justification=raggedright,singlelinecheck=false}
\caption{\textbf{Representative rollouts after \ratstage{} at 2B and 14B scale.} Rows: source video, DreamDojo, and CD-LAM at 2B and 14B; all model rows start from the same initial frame and receive the same robot action sequence.  DreamDojo's arm pose drifts from the ground-truth trajectory, and scaling to 14B does not repair the drift, whereas CD-LAM tracks the commanded motion at both scales. Scene appearance stays comparable, so the separation is in action following rather than visual quality. Colored overlays are row markers distinguishing the models, not model outputs.}
\label{fig:cross_recon}\label{fig:crossrecon}
\end{minipage}
\end{figure}

Table~\ref{tab:posttrain-main} shows that CD-LAM also improves action following after robot-action adaptation. Under robot actions, CD-LAM reduces mean FDCE by 35\% at 2B (12.63 to 8.24) and 30\% at 14B (11.11 to 7.73), while improving median FDCE and all three visual-fidelity metrics. At 14B, CD-LAM achieves the best result in every column of the table, whereas baseline scaling is mixed: FDCE$_{\mathrm{med}}$ \emph{worsens} from 8.15 to 8.98 even as PSNR improves. Figure~\ref{fig:posttrain-panels} further shows that the gains span action categories and shift the rollout distribution toward lower FDCE and higher PSNR. Figure~\ref{fig:appendix-target-and-action} provides the full per-action breakdown and an additional target-action transfer example, while Figure~\ref{fig:cross_recon} shows representative rollouts at both scales. Under zero-input conditioning \(\mathrm{do}(u_t=0)\), residual FDCE decreases from 10.71 to 5.03 at 2B and from 9.36 to 2.18 at 14B. Zero FDCE corresponds to the static reference used in this probe. Under target-action transfer \(\mathrm{do}(u_t=u_t^{\mathrm{tar}})\), FDCE improves from 24.36 to 22.55 and from 24.82 to 21.11. Together, these interventions show that CD-LAM improves action following under both recorded and transferred actions while suppressing residual motion under zero-action conditioning.

\subsection{\Ratstage{} Efficiency and Debiasing Data Scaling}

\label{sec:exp-scale}\label{sec:scale-results}\label{sec:disagree}\label{sec:lessmore}

\begin{figure}[H]
\captionsetup{font=footnotesize,skip=5pt,justification=raggedright,singlelinecheck=false}
\noindent\begin{minipage}[t]{.49\linewidth}\vspace{0pt}
\centering
\centering
\includegraphics[width=\linewidth]{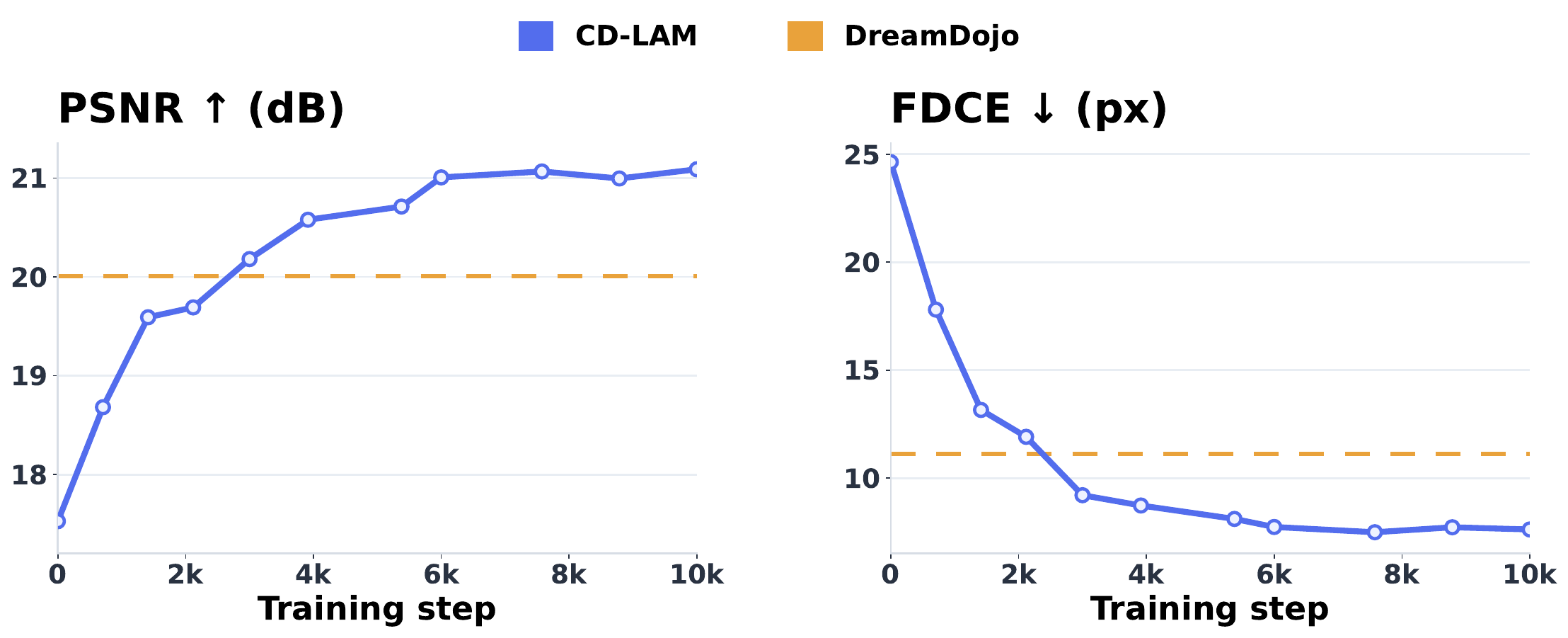}
\captionsetup{justification=raggedright,singlelinecheck=false}
\captionsetup{position=below,skip=6pt}
\caption{\textbf{Robot adaptation efficiency.} 14B PSNR and FDCE over training steps. Dashed lines mark DreamDojo at 50k updates.}
\label{fig:steps}
\end{minipage}\hfill\begin{minipage}[t]{.48\linewidth}\vspace{0pt}
\centering
\captionof{table}{\textbf{Debiasing-data scaling at 2B.} All tiers use 1k Stage-1 updates and differ only in video hours. DreamDojo is the 50k-update robot adaptation reference.}
\label{tab:scaling}
\fontsize{8}{10}\selectfont
\setlength{\tabcolsep}{2pt}
\begin{tabular*}{\linewidth}{@{}l@{\extracolsep{\fill}}ccc@{}}
\toprule
Model & PSNR\,$\uparrow$ & FDCE$_{\text{mean}}$\,$\downarrow$ & FDCE$_{\text{med}}$\,$\downarrow$ \\
\midrule
DreamDojo       & $19.85$ & $12.63$ & $8.15$ \\
\method{}-1h    & $20.54$ & $8.91$  & $6.88$ \\
\method{}-10h   & $20.61$ & $8.87$  & $6.23$ \\
\method{}-100h  & $20.60$ & $8.24$  & $6.75$ \\
\method{}-1000h & $\mathbf{20.64}$ & $\mathbf{7.97}$ & $\mathbf{6.12}$ \\
\bottomrule
\end{tabular*}
\end{minipage}
\end{figure}

Figure~\ref{fig:steps} shows that at 14B, CD-LAM reaches the DreamDojo reference within roughly 3k updates on FDCE and 4k updates on PSNR (more than 12\(\times\) fewer than the reference's 50k updates), and clearly surpasses it on both metrics by the 6k checkpoint in Table~\ref{tab:posttrain-main}. These results suggest that correcting the latent action representation before robot-action adaptation reduces the number of adaptation updates needed to reach the reference performance. Table~\ref{tab:scaling} evaluates the data side of the same effect. All tiers share the same 1k-step Stage-1 budget, so the comparison isolates data quantity at fixed compute. Even the 1h CD-LAM tier reduces FDCE$_{\mathrm{mean}}$ from 12.63 to 8.91, and the 1000h tier further lowers it to 7.97. The 1h tier captures about 80\% of the 1000h tier's FDCE improvement: limited debiasing data yields most of the gain, while more data refines action consistency.

\subsection{Objective Ablation}
\label{sec:objective-ablation-results}\label{sec:ablation}
\begin{table}[!htbp]
\centering
\captionsetup{font=small,position=above,justification=raggedright,singlelinecheck=false,skip=6pt}
\caption{\textbf{CD-LAM objective ablation.} Each row omits the indicated LAM loss term. FG denotes FG-PSNR and $Cam.x/y$ are camera-shift responses.}
\label{tab:objective-ablation}
{\footnotesize
\setlength{\tabcolsep}{6pt}
\renewcommand{\arraystretch}{1.08}
\begin{tabular}{@{}lccccc@{}}
\toprule
Model & PSNR $\uparrow$ & FG $\uparrow$ & $Cam.x\downarrow$ & $Cam.y\downarrow$ & FDCE $\downarrow$ \\
\midrule
Full & \textbf{29.64} & \textbf{26.57} & \textbf{.133} & .101 & 17.23 \\
w/o $\mathcal L_{\mathrm{emb}}$ & 29.53 & 26.26 & .134 & \textbf{.100} & 18.40 \\
w/o $\mathcal L_{\mathrm{ctr}}$ & 29.64 & 26.57 & .134 & .101 & 19.07 \\
w/o $\mathcal L_{\mathrm{cal}}$ & 29.31 & 26.48 & .637 & .637 & \textbf{16.10} \\
\bottomrule
\end{tabular}\par}
\end{table}

Table~\ref{tab:objective-ablation} reports a separate ablation setup: PSNR and FG-PSNR evaluate Stage-2 rollouts, camera-shift norms probe the LAM, and FDCE evaluates Stage-3 rollouts. The $L_{\mathrm{emb}}$ ablation lowers FG-PSNR by 0.31 dB and increases FDCE by 1.17 px. Removing $L_{\mathrm{ctr}}$ leaves the reconstruction and camera-shift scores nearly unchanged but increases FDCE by 1.84 px. Removing the combined latent-space calibration term $L_{\mathrm{cal}}$ increases horizontal/vertical shifted-pair norms from 0.133/0.101 to 0.637/0.637 and lowers PSNR by 0.33 dB, while improving FDCE from 17.23 to 16.10. This reveals a trade-off: $L_{\mathrm{cal}}$ improves latent-space calibration and visual fidelity but does not uniformly improve downstream motion metrics, while the full model still achieves lower FDCE than the $L_{\mathrm{emb}}$ and $L_{\mathrm{ctr}}$ ablations.

\subsection{Efficient Robot-Action Adaptation of a Pretrained Video Backbone}
\label{sec:ltx-enabling}
\begin{table*}[!hbpt]
\centering
\captionsetup{font=footnotesize,position=above,skip=3pt}
\caption{\textbf{Efficient adaptation on LTX-2.3.}
Training exposure and robot-action rollout performance relative to DreamDojo-14B.}
\label{tab:ltx-enabling}

{\fontsize{7.5}{8.5}\selectfont
\setlength{\tabcolsep}{4pt}
\renewcommand{\arraystretch}{1.05}

\begin{adjustbox}{max width=\textwidth}
\begin{tabular}{lccc|ccc|ccc}
\toprule
& \multicolumn{3}{c|}{Latent-action pretraining}
& \multicolumn{3}{c|}{Robot-action adaptation}
& \multicolumn{3}{c}{Robot-action rollout} \\
\cmidrule(lr){2-4}
\cmidrule(lr){5-7}
\cmidrule(lr){8-10}

Model
& GPUs & Updates & \shortstack{Exposure\\Samples}
& GPUs & Updates & \shortstack{Exposure\\Samples}
& PSNR $\uparrow$
& FDCE$_{\mathrm{mean}}$ $\downarrow$
& FDCE$_{\mathrm{med}}$ $\downarrow$ \\
\midrule

DreamDojo 14B
& 256 H100 & 140k & 143.360M
& 128 H100 & 50k & 25.600M
& 20.01 & 11.11 & 8.98 \\

CD-LAM + LTX-2.3 22B
& 8  H200 & 60k & 3.840M
& 7  H200 & 28k & 1.568M
& \textbf{20.38} & 11.75 & \textbf{7.82} \\

\bottomrule
\end{tabular}
\end{adjustbox}
}
\end{table*}

We further test whether CD-LAM can efficiently adapt a generic pretrained video backbone into an embodied action-conditioned world model. Starting from LTX-2.3 22B, we perform only 60k latent-action pretraining updates, followed by 28k robot-action adaptation updates. This corresponds to 37.3 $\times$ fewer latent-action training sample exposures and 16.3 $\times$ fewer robot-action adaptation exposures than DreamDojo-14B. Despite this substantially reduced training exposure, the adapted LTX-2.3 achieves comparable robot-action rollout quality, with higher PSNR (20.38 vs.\ 20.01) and lower median FDCE (7.82 vs.\ 8.98). These results show that CD-LAM supports efficient adaptation of a pretrained video backbone to robot-action conditioning, achieving comparable rollout quality with substantially fewer training cost than DreamDojo-14B.

\subsection{Debiased Latent Actions as a Policy Pretraining Signal}
\label{sec:exp-vla}
\begin{figure}[!htbp]
\captionsetup{font=footnotesize,skip=5pt,justification=raggedright,singlelinecheck=false}
\noindent\begin{minipage}[t]{.51\linewidth}\vspace{0pt}
\centering
\includegraphics[width=\linewidth]{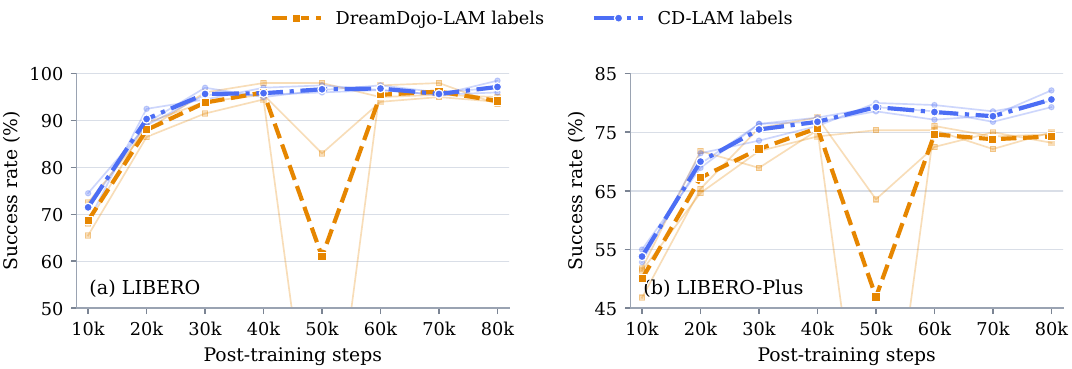}
\caption{\textbf{Policy learning curves.} LIBERO (left) and LIBERO-Plus (right). Means, individual seeds, and min--max bands over three paired seeds; fixed subsets of 200 and 280 episodes. Full-benchmark comparisons use the 80k endpoints.}
\label{fig:vla-steps}
\end{minipage}\hfill\begin{minipage}[t]{.47\linewidth}\vspace{0pt}
\centering
\centering
\captionsetup{font=footnotesize,justification=raggedright,singlelinecheck=false,skip=6pt}
\captionof{table}{\textbf{X-VLA policy success (\%).} CD-LAM improves over DreamDojo LAM on LIBERO, LIBERO-Plus, and RoboTwin C2R.}
\label{tab:vla-main}
{\fontsize{7.5}{9.5}\selectfont
\setlength{\tabcolsep}{2pt}
\renewcommand{\arraystretch}{1.08}
\begin{tabular*}{\linewidth}{@{}l@{\extracolsep{\fill}}cc|cc@{}}
\toprule
\multirow{2}{*}{Training} & \multirow{2}{*}{LIBERO} & \multirow{2}{*}{\shortstack{LIBERO-\\Plus}} & \multicolumn{2}{c}{RoboTwin C2R} \\
& & & Clean & Random \\
\midrule
X-VLA & 98.05 & 69.74 & 85.67 & 25.00 \\
\arrayrulecolor{black!25}\midrule\arrayrulecolor{black}
DreamDojo LAM & 95.72 & 75.69 & 84.00 & 28.83 \\
CD-LAM & \textbf{96.30} & \textbf{78.25} & \textbf{86.67} & \textbf{33.83} \\
\midrule
$\Delta$ (pp) & +0.58 & +2.56 & +2.67 & +5.00 \\
\bottomrule
\end{tabular*}}
\end{minipage}
\end{figure}

Finally, we compare X-VLA policies trained with latent actions from either DreamDojo LAM or CD-LAM, using matched recipes within each benchmark (Table~\ref{tab:vla-main}). Across three paired post-training seeds, CD-LAM improves LIBERO success from 95.72\% to 96.30\% and LIBERO-Plus overall success from 75.69\% to 78.25\%, with positive mean gains across all seven perturbation categories. On RoboTwin~2.0, we evaluate C2R robustness on a predefined 12-task subset comprising ten eligible tasks with the largest public X-VLA clean-to-random gaps and two anchor tasks to assess policy robustness after latent-action pretraining. With  CD-LAM, randomized-scene success increases from 28.8\% to 33.8\%, with comparable clean performance (84.0\% vs.\ 86.7\%). These results show that CD-LAM also improves downstream policy success. Full protocols, task-selection criteria, and task-level results are provided in Appendix~\ref{app:vla}.
Figure~\ref{fig:vla-steps} shows the three paired post-training trajectories on fixed monitoring subsets; the full-benchmark comparisons above use the fixed 80k endpoints.

\FloatBarrier

\section{Conclusion}\label{sec:conclusion}
We presented CD-LAM, a causally motivated framework that improves world model controllability by reducing visual bias in latent action representations. Across 2B and 14B backbones, CD-LAM improves action following, visual fidelity, and adaptation efficiency, with benefits persisting from latent action conditioning to executable robot controls. Experiments with LTX-2.3 and X-VLA further demonstrate its applicability to video model adaptation and policy learning. Together, these findings identify upstream representation bias as a practical obstacle to downstream control and show that correcting it yields benefits across models and tasks. By correcting bias in latent actions, CD-LAM offers a practical path toward more controllable world models and more robust embodied policies.

\FloatBarrier

\clearpage
\bibliography{references}
\bibliographystyle{iclr2027_conference}
\clearpage
\appendix
\providecommand{\FDCE}{\operatorname{FDCE}}

\counterwithin{equation}{section}
\counterwithin{figure}{section}
\counterwithin{table}{section}
\renewcommand{\theHequation}{appendix.\thesection.\arabic{equation}}
\renewcommand{\theHfigure}{appendix.\thesection.\arabic{figure}}
\renewcommand{\theHtable}{appendix.\thesection.\arabic{table}}

\section{Related Work}\label{sec:related}
\indent{\textbf{\textit{Latent Action Learning from Video.}}}
Latent action learning builds on observation-only imitation~\citep{edwards2019ilpo} and video-based pretraining with inferred action labels~\citep{baker2022vpt}. Genie~\citep{bruce2024genie} learns discrete latent controls for interactive generation, while LAPO~\citep{schmidt2024lapo} uses inferred actions to bootstrap policies and shows that improved reconstruction need not yield better policy learning. LAPA~\citep{ye2024lapa} and Moto~\citep{chen2024moto} use quantized transition representations for policy pretraining: LAPA replaces its latent prediction head during robot-action fine-tuning, whereas Moto retains motion-token supervision. These works establish the value of video-derived action targets, while leaving different requirements on the representations ultimately used for control.

\indent{\textbf{\textit{World Models and Latent-Action Conditioning.}}}
World models predict environment dynamics for planning and control~\citep{ha2018worldmodels,hafner2023dreamerv3}, with UniSim~\citep{yang2024unisim} and Cosmos~\citep{nvidia2025cosmos} scaling generative modeling to diverse video data. Action-conditioned architectures~\citep{acwmphys2026,wu2024ivideogpt} and robot datasets spanning diverse tasks and scenes~\citep{oxe2023,khazatsky2024droid} support learning from executable controls. To exploit action-unlabeled video, IGOR~\citep{chen2024igor} uses image-goal representations for both policy learning and video generation, while AdaWorld~\citep{gao2025adaworld} and DreamDojo~\citep{dreamdojo2026} use continuous latent actions for world-model pretraining and subsequent adaptation. CoLA-World~\citep{colaworld2025} couples quantized latent-action learning with a pretrained world model and analyzes distribution mismatch and codebook degeneration during real-action adaptation. CLAW~\citep{ayalew2026claw} jointly learns continuous actions and a diffusion world model, adversarially suppressing future-observation leakage. These approaches improve latent--world-model learning, but suppressing future appearance does not by itself exclude exogenous motion, and codebook stability does not determine which physical factors the codes represent.

\indent{\textbf{\textit{Action-Relevant and Structured Latent Representations.}}}
\citet{zhang2025whatdolams} analyze how reconstruction-driven learning can encode exogenous variation and observation-dependent shortcuts. LAOM~\citep{nikulin2025latentdistractors} studies distractors and shows that action decodability and usefulness as a policy target are distinct considerations. More recent methods impose explicit structure. ConLA~\citep{conla2026} combines category-based contrastive learning and temporal-reversal constraints in a quantized RGB-reconstruction LAM, while MVP-LAM~\citep{mvplam2026} uses cross-view reconstruction of frozen DINO features. Category agreement and cross-view consistency constrain semantic organization and viewpoint dependence, but do not by themselves separate all embodiment-controlled changes from other predictable variation. AC-LAM~\citep{aclam2026} imposes additive composition and evaluates identity, inverse consistency, and normalized zero-transition responses; these algebraic properties provide useful structure without uniquely determining the represented physical factors. MaskLAM~\citep{fechner2026masklam} restricts reconstruction to agent regions to reduce distractor encoding. DiLA~\citep{zhang2026dila} learns continuous actions through content--structure disentanglement and dynamics prediction in DINO feature space, supporting RGB generation and visual planning. Its separation of appearance from structural change leaves open how reliably transferred motion is assigned to the intended entity. These distinctions motivate evaluating latent actions according to their downstream use as embodied control conditions.

\indent{\textbf{\textit{Causal Perspectives on Bias Propagation.}}}
Invariant prediction~\citep{arjovsky2019irm}, shortcut learning~\citep{geirhos2020shortcut}, and causal confusion in imitation learning~\citep{dehaan2019causalconfusion} motivate examining which information a predictor uses and how that dependence affects its behavior. We investigate a specific mechanism in LAM-based ACWMs. CoCo~\citep{shi2026coco} addresses statistical shortcuts through counterfactual consistency constraints on action-conditioned world models. Source-specific visual information can be useful through the observation pathway yet also enter the latent action. If the world model relies on this information through its action condition, alignment with robot commands must accommodate variation that those commands cannot generally determine. CD-LAM targets this bias at its source through embodiment-centric reconstruction, action-centric contrastive learning, and latent-space calibration. We connect encoder diagnostics with action interventions and conditioned rollouts before and after \ratstage{}, evaluating how upstream representation correction affects controllability and whether its benefits persist under executable robot controls. Our study focuses on correcting the upstream latent-action representation and evaluating its effects before and after robot-action adaptation.

\indent{\textbf{\textit{Evaluating Controllability and Embodied Motion.}}}
Visual-fidelity metrics (\eg{} PSNR and FVD~\citep{unterthiner2018fvd}) do not by themselves establish whether a rollout follows its supplied action condition. Genie measures action sensitivity using $\Delta$PSNR~\citep{bruce2024genie}. MotionCtrl evaluates object-motion control with ObjMC~\citep{motionctrl2024}, while MotionPro measures trajectory alignment using mean-distance metrics, including MD-Vid computed with CoTracker~\citep{motionpro2025,cotracker2024}. Our \fdce{} metric compares foreground displacements between generated and reference rollouts, using SAM3 masks~\citep{carion2025sam3} and tracks from CoWTracker~\citep{lai2026cowtrackertrackingwarpinginstead}. We interpret this motion measure jointly with visual fidelity, latent diagnostics, and zero-action and target-action interventions to assess the downstream effects of latent-action debiasing.

\section{Training and Evaluation Details}
\label{app:setup}
This appendix specifies the main DreamDojo-based experiments. The LTX and X-VLA protocols are given separately in Appendices~\ref{app:ltx} and~\ref{app:vla}.

\subsection{Training Data and Evaluation Sets}
\label{app:data}
\textbf{Training data.} Unless otherwise stated, Stage 1 uses 100h of debiasing video drawn from DreamDojo's pretraining data. The data-scaling study varies this amount over 1h, 10h, 100h, and 1000h while keeping the Stage-1 budget at 1k updates. Stage 2 compares CD-LAM with continued DreamDojo training on the same video subset and with matched update counts. Stage 3 uses approximately 300h of paired AgiBot video and executable actions~\citep{agibotworldcontributors2025agibotworldcolosseolargescale}.

\textbf{Evaluation sets.} Stage-2 latent-action rollouts are evaluated on 300 EgoDex clips~\citep{hoque2026egodexlearningdexterousmanipulation}, an egocentric human-manipulation corpus. Stage-3 robot-action rollouts and interventions are evaluated on 300 held-out AgiBot clips drawn from distinct episodes. Metric definitions and validity filtering are given in Appendix~\ref{app:metrics}.

\subsection{Stage-specific Training Configuration}
\label{app:training}
The three stages follow Section~\ref{sec:method-bridge}. All main-study runs use 128 H100 GPUs. Table~\ref{tab:training-budgets} summarizes the update budgets and per-GPU batch sizes. Stage-3 entries identify the checkpoints used for the main results.

\begin{table}[!htbp]
\centering\small
\caption{\textbf{Main-study training budgets.} The same debiased LAM serves both ACWM scales. Stage-3 counts exclude the preceding Stage-1 and Stage-2 updates. The released DreamDojo Stage-3 reference uses 50k updates; Stage-3 batches match this reference.}
\label{tab:training-budgets}
\setlength{\tabcolsep}{5pt}
\renewcommand{\arraystretch}{1.05}
\begin{tabular*}{\linewidth}{@{}l@{\extracolsep{\fill}}lll@{}}
\toprule
Stage & Model / backbone & Optimizer updates & Batch / GPU \\
\midrule
1: LAM debiasing & Shared LAM & 1k & 48 \\
2: ACWM fine-tuning & 2B & 2k & 12 \\
2: ACWM fine-tuning & 14B & 2k & 2 \\
3: Robot-action adaptation & 2B & 3k checkpoint & Matched \\
3: Robot-action adaptation & 14B & 6k checkpoint & Matched \\
\bottomrule
\end{tabular*}
\end{table}

\textbf{Objectives and optimization.} Stage-3 batch sizes and optimizer settings match the DreamDojo reference (Section~\ref{app:comparison}). Table~\ref{tab:loss} gives the Stage-1 loss weights, schedules, and remaining reported LAM settings. Bridge fitting has its own optimizer and batch size (Section~\ref{app:bridge}); these are distinct from ACWM training.

\begin{table}[!htbp]

\caption{\textbf{\method{} objective hyperparameters.} Loss weights and schedules behind Equations~(\ref{eq:cdlam-total}), (\ref{eq:emb-rec}), (\ref{eq:contrast-compact}), and (\ref{eq:action-norm-compact}), in the notation of Section~\ref{sec:method}. Pair weighting for Equation~(\ref{eq:contrast-compact}) is specified in the released configuration. Stage-specific update budgets are summarized in Table~\ref{tab:training-budgets}.}
\label{tab:loss}
\centering\footnotesize
\setlength{\tabcolsep}{5pt}\renewcommand{\arraystretch}{1.06}
\begin{tabular}{@{}lll@{}}
\toprule
Term & Hyperparameter & Value \\
\midrule
\multicolumn{3}{@{}l}{\emph{Embodiment-centric Reconstruction} $\mathcal{L}_{\mathrm{emb}}$ (foreground / background)}\\
foreground recon weight     & $\alpha_{\mathrm{fg}}$   & $1.0$ \\
background recon weight      & $\alpha_{\mathrm{bg}}$    & $0.02$ \\
\midrule
\multicolumn{3}{@{}l}{\emph{Action-centric Contrastive Learning} $\mathcal{L}_{\mathrm{ctr}}$}\\
projection head             & $r_\omega$                & $32\!\to\!128\!\to\!32$ \\
log-scale / bias init & $\log\tau,\,b$ & $2.30,\ -3.0$ \\
coefficient schedule        & $\lambda_{\mathrm{ctr}}(k)$ & $0\to0.020\,(\text{steps }150\text{--}400)\to0.005\,(\cos,\ \text{by }1000)$ \\
\midrule
\multicolumn{3}{@{}l}{\emph{Latent Space Calibration} $\mathcal{L}_{\mathrm{cal}}$ (free-bit KL + zero-transition calibration)}\\
KL free bits                & $\kappa$                & $0.5$ \\
KL weight (ramped)          & $\beta_{\mathrm{kl}}(k)$  & $10^{-4}\!\to\!10^{-3}$ by step $500$ \\
zero-transition weight      & $\lambda_{\mathrm{zero}}$ & $1.0$ \\
zero-transition margin      & $m_{\mathrm{zero}}$         & $0.05$ \\
zero-transition constant & $\epsilon$ & $10^{-8}$ \\
running RMS $s_\Delta$ & --- & EMA of squared norm, update weight $0.05$ \\
\midrule
\multicolumn{3}{@{}l}{\emph{Latent / optimization}}\\
latent-action dim          & ---                     & $32$ \\
frame size / precision     & ---                     & $240\times320$ / bf16 \\
weight decay / grad clip   & ---                     & $10^{-4}$ / $1.0$ \\
initialization             & ---                     & DreamDojo LAM checkpoint \\
\bottomrule
\end{tabular}
\end{table}

\subsection{Baseline and Efficiency Comparison}
\label{app:comparison}
For Stage 2, CD-LAM and the continued-training DreamDojo baseline use the same video subset and matched update counts. For Stage 3, the main results compare the adapted CD-LAM pipeline with DreamDojo's released robot-action-adapted checkpoints under the same evaluation protocol.

In the robot-action adaptation efficiency comparison, step counts denote optimizer updates under the aligned protocol: both models use the same adaptation data, batch size, and optimizer settings. The 50k-step reference in Figure~\ref{fig:overall}(c) and the dashed lines in Figure~\ref{fig:steps} refer to the released DreamDojo AgiBot checkpoint, trained for 50k updates on 128 GPUs with the same per-GPU batch as our adaptation runs. The reported update-count reduction concerns Stage 3; it excludes the preceding 1k LAM debiasing and 2k ACWM fine-tuning updates and is not an end-to-end wall-clock speedup. The LTX comparison uses different backbones and hardware and is not compute-normalized (Appendix~\ref{app:ltx}).

\subsection{Robot-action Bridge}
\label{app:bridge}
The bridge \(g_\eta\) and the auxiliary readout \(h\) are three-layer MLPs (Linear--GELU--Linear--GELU--Linear, width 256). \(g_\eta\) maps the full 22-D action vector \(u^{\mathrm{in}}\) to the 32-D latent, and \(h\) maps a latent back to \(u^{\mathrm{act}}\), the 15 action dimensions that vary in the training split; \(u\) and \(z\) are standardized per dimension with training-split statistics. They are fitted by minimizing \(\lambda_{\mathrm{enc}}\|g_\eta(u^{\mathrm{in}})-z\|_2^2+\lambda_{\mathrm{cyc}}\|h(g_\eta(u^{\mathrm{in}}))-u^{\mathrm{act}}\|_2^2+\lambda_{\mathrm{dec}}\|h(z)-u^{\mathrm{act}}\|_2^2\) with weights \(1.0/3.0/0.3\), using Adam (learning rate \(10^{-3}\), weight decay \(10^{-5}\)), batch 2{,}048 and 300 epochs on 10{,}560 training and 1{,}440 validation pairs; the LAM is frozen and its latents are pre-computed. During \ratstage{} the bridge is frozen as well, so the video loss does not back-propagate into \(g_\eta\).

\subsection{Intervention Protocol}
\label{app:intervention}
Interventions follow Section~\ref{sec:observation}: the observation history is held fixed and only the conditioning action is replaced. Under a \emph{zero action}, \(\mathrm{do}(u_t=0)\), embodiment motion should be suppressed. Under \emph{target-action transfer}, the conditioning action is replaced with one encoded from a different target video, and the rollout should reflect the direction and magnitude of the transferred target action under the fixed source context. For latent actions this means \(\mathrm{do}(z_t=z_t^{\mathrm{tar}})\) with \(z_t^{\mathrm{tar}}=\mu_\phi(o_t^{\mathrm{tar}},o_{t+1}^{\mathrm{tar}})\); for robot actions, \(\mathrm{do}(u_t=u_t^{\mathrm{tar}})\) with \(u_t^{\mathrm{tar}}\) the target video's recorded action.

\FloatBarrier

\section{Metric Details}
\label{app:metrics}

\noindent\textbf{PSNR Reporting.}
We report PSNR as the visual-fidelity metric, computed on full frames in dB. For images normalized to \([0,1]\),
\begin{equation}
\mathrm{PSNR}(x,\hat{x})
=
10\log_{10}
\frac{1}{\operatorname{MSE}(x,\hat{x})}.
\label{eq:psnr-mse}
\end{equation}
Figure~\ref{fig:overall}(b) reports absolute PSNR for each setting, with CD-LAM gains annotated in dB. Because PSNR is logarithmic, a fixed dB gain corresponds to a multiplicative MSE reduction: the \(+1.0\)\,dB gain at 14B after \ratstage{} is a \(20.6\%\) reduction in MSE (\(10^{-1/10}\approx0.794\)).

\noindent\textbf{\fdce{} Definition.}
\fdce{} measures foreground action following by comparing foreground displacement tracks rather than raw pixels. It draws on trajectory-based evaluation for controllable video generation~\citep{motionctrl2024,motionpro2025}, where generated trajectories are compared with prescribed motion controls. Our adaptation compares foreground displacement tracks extracted from generated and reference videos. Foreground masks select embodiment and interacted-object regions using SAM3~\citep{carion2025sam3}, and point tracks are computed only within valid foreground regions using CoWTracker~\citep{lai2026cowtrackertrackingwarpinginstead}.

For a reference foreground point \(p_j^{s}\) and a generated foreground point \(\hat{p}_i^{s}\) at rollout step \(s\), we define displacement vectors relative to the initial frame as
\begin{equation}
a_j^{s}
=
p_j^{s} - p_j^0,
\qquad
\hat{a}_i^{s}
=
\hat{p}_i^{s} - \hat{p}_i^0 .
\label{eq:fdce-displacement}
\end{equation}

For rollout step \(s\), let \(\hat{\mathcal A}^{s}=\{\hat a_i^{s}\}_{i=1}^{N_g^{s}}\) and \(\mathcal A^{s}=\{a_j^{s}\}_{j=1}^{N_r^{s}}\) be the displacements of the \(N_g^{s}\) generated and \(N_r^{s}\) reference foreground tracks that are valid at that step. The per-step symmetric Chamfer distance is
\begin{equation}
d^{s}
=
\frac{1}{2N_g^{s}}\sum_{i=1}^{N_g^{s}}\min_{j}\left\|\hat a_i^{s}-a_j^{s}\right\|_2
+
\frac{1}{2N_r^{s}}\sum_{j=1}^{N_r^{s}}\min_{i}\left\|\hat a_i^{s}-a_j^{s}\right\|_2 ,
\label{eq:fdce-track-cost}
\end{equation}
and \fdce{} averages it over the retained rollout steps \(\mathcal S_{\mathrm{valid}}\) (steps with fewer than four valid points on either side are dropped, see below):
\begin{equation}
\FDCE(\hat{o},o)
=
\frac{1}{|\mathcal S_{\mathrm{valid}}|}\sum_{s\in\mathcal S_{\mathrm{valid}}} d^{s} .
\label{eq:fdce}
\end{equation}
In our evaluation, we sample up to 16 valid foreground anchors per rollout pair and average the bidirectional nearest-neighbor distances. Lower \fdce{} indicates that the generated rollout induces foreground displacement closer to the reference action. Figure~\ref{fig:psnr-fdce} contrasts \fdce{} with PSNR on the same rollouts.

\noindent\textbf{Robustness Properties.}
  Anchors are seeded inside an eroded foreground mask. Tracks are linearly interpolated over short invisible spans and Savitzky--Golay smoothed before scoring; at step \(s\) only points visible at both step 0 and step \(s\) enter \(\hat{\mathcal A}^{s}\) and \(\mathcal A^{s}\), steps with fewer than four valid points on either side are skipped, and a rollout pair with fewer than two valid steps is marked invalid and excluded from all means (596 of the 600 arm--pairs of the 300-clip evaluation are valid). The symmetric Chamfer form in Equation~(\ref{eq:fdce-track-cost}) scores displacement geometry rather than matched point counts, so the metric is robust to differing numbers of valid tracks.  Tracking can be unreliable under heavy occlusion or fast motion, which can affect \fdce{} estimates.

\noindent\textbf{Reporting Conventions.}
\fdce{} is reported in pixels at the evaluation resolution.  The mean is sensitive to large outliers while the median reflects typical behavior; we report both. Pixel metrics (PSNR, SSIM, LPIPS) are computed on full frames against the reference rollout, and FG-PSNR restricts the same computation to the foreground mask.

\noindent\textbf{LAM Diagnostic Definitions.}
The three diagnostics of Table~\ref{tab:latent-audit} are computed from the posterior mean \(\mu_\phi\) on the audit split and reported as medians (camera shifts also as means; shortcut leakage is a difference of means) over evaluation pairs.
The \emph{zero-transition} and \emph{camera-shift} responses are computed from a duplicated pair and a synthetically shifted pair, respectively, normalized by the typical latent norm of ordinary transitions:
\begin{equation}
R_{\mathrm{zero}}
=
\frac{\|\mu_\phi(o_t,o_t)\|_2}{D},
\qquad
R_{\mathrm{shift}}
=
\frac{\|\mu_\phi(o_t,T_{3}(o_t))\|_2}{D},
\label{eq:diag-resp}
\end{equation}
where \(D=\operatorname{RMS}\big(\|\mu_\phi(o_t,o_{t+1})\|_2\big)+\epsilon\) over the audit split and \(T_{3}\) translates the frame horizontally or vertically by 3 pixels.
The \emph{shortcut leakage} is the cosine gap
\begin{equation}
\begin{aligned}
L_{\mathrm{shortcut}}
={}&
\mathbb{E}\big[\cos(z_i,z_j)\,\big|\,\text{same episode, diff. primitive}\big]\\
&-
\mathbb{E}\big[\cos(z_i,z_j)\,\big|\,\text{diff.\ episode, same primitive}\big],
\end{aligned}
\label{eq:diag-shortcut}
\end{equation}
where pairs are drawn from the audit split using the coarse primitive labels of Appendix~\ref{app:verbs}; the second term alone  is the same-primitive cosine similarity reported in the main text (0.132 vs.\ 0.131). 

\noindent\textbf{Reading Table~\ref{tab:latent-audit}.}
All three diagnostics feed the encoder a change that is not an embodied action, so an encoder whose latent carries only embodiment dynamics should respond weakly to all of them. Table~\ref{tab:latent-audit-full} expands Table~\ref{tab:latent-audit} with the input of each diagnostic and its ideal value.
\begin{itemize}
\item \emph{Zero-transition.} A duplicated pair \((o_t,o_t)\) contains no motion, so its latent should lie near zero. \(R_{\mathrm{zero}}\) expresses its norm as a fraction of a typical transition latent. The DreamDojo LAM gives 0.527, \ie{} a static input is encoded as roughly half of a typical action, whereas \method{} gives 0.043. The absolute norm \(\|\mu_\phi(o_t,o_t)\|_2\) (3.119 vs.\ 0.226) is reported alongside to show that the relative drop does not come from a larger normalizer \(D\).
\item \emph{Camera shift.} \(T_3\) moves the whole frame by 3 pixels, which mimics a small camera motion while no hand or object moves. The DreamDojo LAM responds to such a shift with more than half the norm of a typical transition (0.53--0.56), and \method{} responds 3.6--8.3\(\times\) more weakly.
\item \emph{Shortcut leakage.} This diagnostic asks what makes two latents similar. For the DreamDojo LAM, pairs from the same episode with different primitives have a mean cosine similarity of 0.283, higher than pairs with the same primitive from different episodes (0.132), so shared visual context outweighs shared action. \method{} reduces this gap from 0.151 to 0.014 while the same-primitive term stays essentially unchanged (0.131). The reduction therefore comes from removing context-driven similarity, not from shrinking all similarities. A value near zero means episode context no longer outweighs action in the latent neighborhood structure.
\end{itemize}

\begin{table}[!htbp]
\centering\small
\caption{\textbf{Expanded LAM encoder audit} (Table~\ref{tab:latent-audit} with inputs and ideal values). Norm responses are relative to \(D\), the RMS latent norm of ordinary transitions, except the absolute norm. Camera-shift entries are mean/median. All other responses are medians, and cosine terms are means over pairs.}
\label{tab:latent-audit-full}
\setlength{\tabcolsep}{4pt}
\resizebox{\linewidth}{!}{%
\begin{tabular}{@{}llccc@{}}
\toprule
Diagnostic & Encoder input & Ideal & DreamDojo LAM & \method{} \\
\midrule
Zero-transition, rel.\ norm \(R_{\mathrm{zero}}\) & duplicated pair \((o_t,o_t)\) & \(\approx0\) & \(0.527\) & \(\mathbf{0.043}\) \\
Zero-transition, abs.\ norm & duplicated pair \((o_t,o_t)\) & \(\approx0\) & \(3.119\) & \(\mathbf{0.226}\) \\
\addlinespace[2pt]
Camera shift, horizontal \(R_{\mathrm{shift}}\) & \((o_t,T_3(o_t))\), 3\,px & \(\approx0\) & \(0.555/0.536\) & \(\mathbf{0.156/0.096}\) \\
Camera shift, vertical \(R_{\mathrm{shift}}\) & \((o_t,T_3(o_t))\), 3\,px & \(\approx0\) & \(0.545/0.529\) & \(\mathbf{0.110/0.064}\) \\
\addlinespace[2pt]
Shortcut leakage \(L_{\mathrm{shortcut}}\) & latent pairs, Equation~(\ref{eq:diag-shortcut}) & \(\le0\) & \(0.151\) & \(\mathbf{0.014}\) \\
\quad same-primitive, diff.\ episode cosine & latent pairs & kept & \(0.132\) & \(0.131\) \\
\bottomrule
\end{tabular}}
\end{table}

\section{Coarse Action-primitive Labels}
\label{app:verbs}
The action-centric contrastive loss (Equation~(\ref{eq:contrast-compact})) uses coarse action-primitive labels rather than executable robot actions. The label space is a 12-way canonical verb set: pick--place, insert--remove, stack--unstack, scoop--dump, open, close, turn on, turn off, wash--rinse, cut, stir, and pour.  Labels come from the task-name and verb fields shipped with the source datasets. A hand-written rule table maps each verb phrase and its aliases (e.g., grab, grasp, take \(\to\) pick--place) onto one of the 12 canonical primitives; clips whose annotation matches no rule remain unlabeled and join no contrastive pair. The labels are verb-level only (no controller states or trajectories) and merely tell \(\mathcal L_{\mathrm{ctr}}\) which transitions plausibly share a primitive.

Of the 68{,}864 transition pairs in the clean-ego index, 25{,}192 (36.6\%) carry a primitive label; unlabeled pairs contribute only to reconstruction and calibration.  The LAM audit split covers all 12 primitives. The label distribution is long-tailed (pick--place dominates), so a typical training batch exposes the contrastive term to an effective 8--10 primitives (exponentiated label entropy). The alias table deliberately merges phrasing- and viewpoint-dependent verb variants, so \(\mathcal L_{\mathrm{ctr}}\) never relies on fine-grained verb semantics.

\section{Additional Results}
\label{app:qual}
\begin{figure}[!tbp]
\centering
\includegraphics[
    width=0.72\textwidth,
    height=0.5\textheight,
    keepaspectratio
]{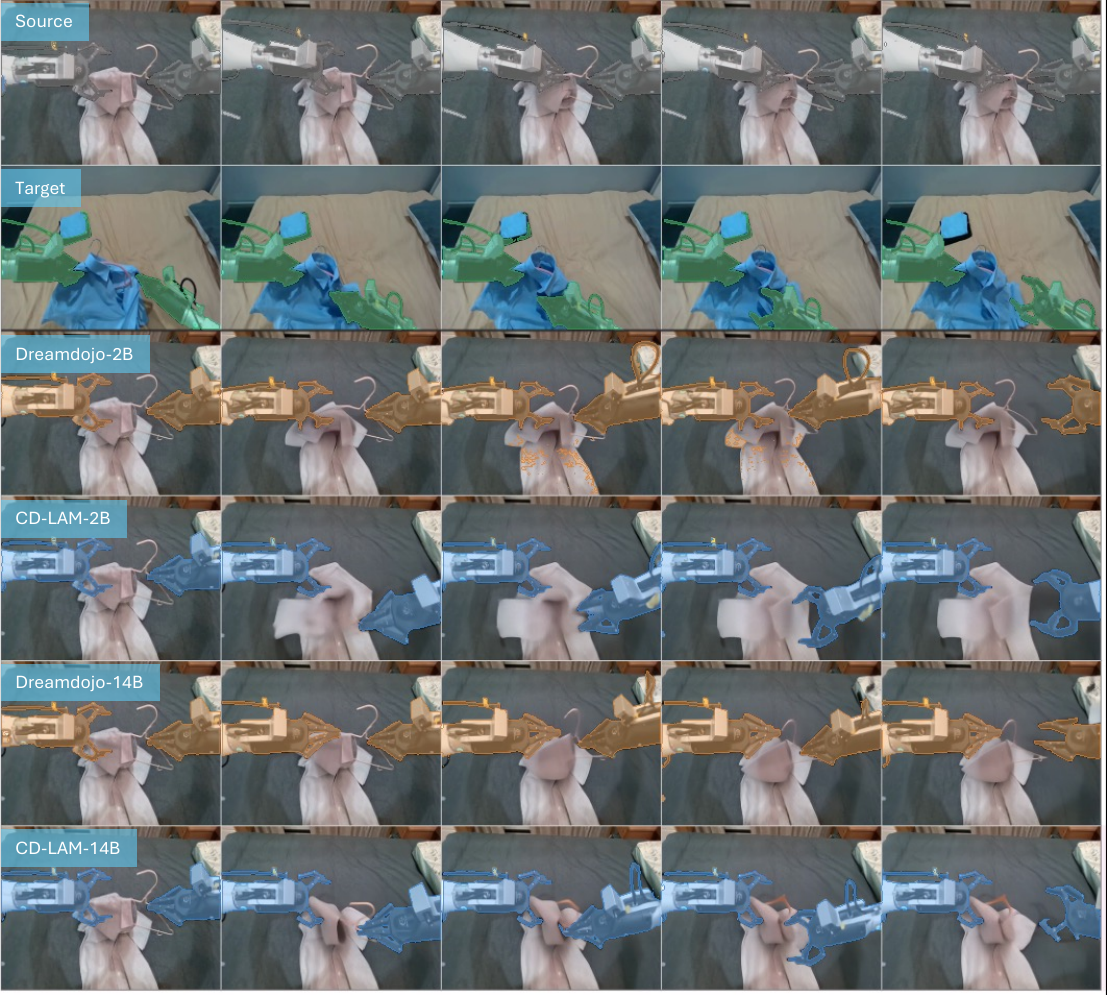}

\vspace{1.2em}

\includegraphics[
    width=0.95\textwidth,
    height=0.4\textheight,
    keepaspectratio
]{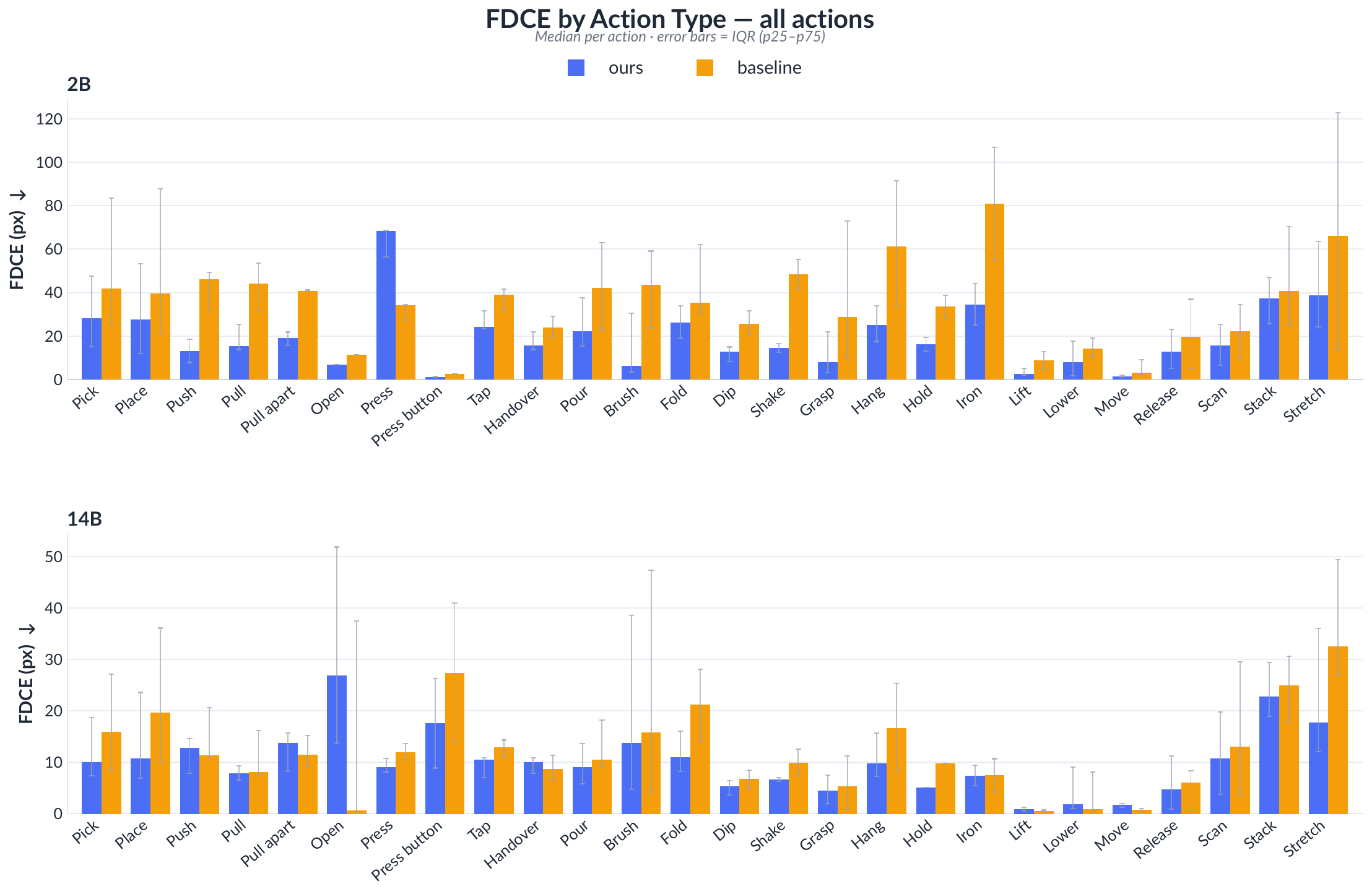}

\vspace{0.4em}

\caption{
\textbf{Additional target-action and per-action results.}
\textbf{Top:} Target-action transfer example. The source-context row provides the fixed visual context, the target-motion row provides the robot action sequence, and rollouts are generated under the same target action. CD-LAM transfers the target motion more reliably at both scales.
\textbf{Bottom:} Per-action \fdce{} breakdown (median; error bars show the inter-quartile range). CD-LAM lowers \fdce{} across most categories.
}
\label{fig:appendix-target-and-action}
\end{figure}

\renewcommand{\floatpagefraction}{0.75}
Figure~\ref{fig:posttrain-panels} breaks down the \ratstage{} results of Section~\ref{sec:exp-posttrain} by action category and rollout distribution, and Figure~\ref{fig:appendix-target-and-action} gives the full per-action breakdown together with an additional target-action transfer example. Figure~\ref{fig:cross_recon} shows representative rollouts after \ratstage{} at both scales. The baseline target-action transfer failure is shown together with the zero-action intervention in Figure~\ref{fig:action-intervention-failures}. In Figure~\ref{fig:transferaction}, the first row shows the source context, and the second row shows the target video that provides the robot action sequence. The third row is the ACWM rollout conditioned on that target action under the fixed source context. The rollout does not reproduce the target embodiment dynamics, indicating target-action misalignment after \ratstage{}.

\section{LTX Adaptation: Implementation Details}
\label{app:ltx}

\paragraph{Integrating CD-LAM into LTX.}
We introduce CD-LAM conditioning into the pretrained LTX-2.3 22B backbone~\citep{hacohen2026ltx2,lightricks2026ltx23} through adaptive layer normalization (AdaLN). Given a 32-dimensional latent action extracted by CD-LAM, an action embedder produces an embedding $e_a$, which is projected and combined with the diffusion-timestep embedding $e_\tau$:
\begin{equation}
c=e_\tau+P_a(e_a).
\label{eq:ltx-conditioning}
\end{equation}
Each DiT block maps $c$ to channel-wise scale, shift, and residual-gate parameters for its attention and feed-forward sublayers. CD-LAM latents therefore condition the backbone through AdaLN modulation rather than additional cross-attention tokens. During subsequent robot-action adaptation, a learned bridge maps 22-dimensional executable actions into the same 32-dimensional conditioning space.

\paragraph{CD-LAM-conditioned midtraining.}
After introducing this conditioning pathway, we midtrain LTX with CD-LAM latent actions for 60k optimizer updates on approximately 20{,}000 hours of egocentric video. This stage uses 8 H200 GPUs, a per-GPU micro-batch size of 2, and gradient accumulation over 4 micro-batches, yielding an effective global batch size of 64 and 3.84M processed examples. The video duration describes the midtraining corpus, while the processed-example count includes repeated sampling.

\paragraph{Robot-action adaptation.}
We subsequently perform full-model robot-action adaptation on the same paired AgiBot video and action data used for DreamDojo adaptation. Adaptation uses 7 H200 GPUs with the same per-GPU micro-batch size and gradient accumulation, yielding an effective global batch size of 56. We evaluate checkpoints at 17k, 23k, and 28k adaptation updates, corresponding to 0.952M, 1.288M, and 1.568M processed examples, respectively; the main text reports the 28k checkpoint. All step counts refer to optimizer updates rather than individual gradient-accumulation micro-steps. Adaptation counts restart after, and exclude, the preceding 60k CD-LAM-conditioned midtraining updates.

\paragraph{Evaluation and additional checkpoints.}
Given an initial frame and a sequence of executable robot actions, the adapted model generates an autoregressive video rollout. We use the same held-out AgiBot evaluation protocol as the main robot-action study. Table~\ref{tab:ltx-full-results} reports all evaluated checkpoints, including LPIPS in addition to the PSNR and FDCE metrics shown in the main text. Metric definitions follow Appendix~\ref{app:metrics}. These results evaluate action-conditioned generation within the AgiBot data domain and do not establish generalization across unseen domains.

\begin{table}[!htbp]
\centering
\small
\caption{\textbf{LTX robot-action adaptation checkpoints.} All checkpoints follow the same 60k-update CD-LAM-conditioned midtraining stage. Processed examples count robot-action adaptation only and include repeated sampling. PSNR is in dB and FDCE in pixels.}
\label{tab:ltx-full-results}
\setlength{\tabcolsep}{4pt}
\renewcommand{\arraystretch}{1.08}
\begin{tabular*}{\linewidth}{@{}c@{\extracolsep{\fill}}ccccc@{}}
\toprule
Updates & Examples & PSNR $\uparrow$ & LPIPS $\downarrow$ & FDCE$_{\mathrm{mean}}$ $\downarrow$ & FDCE$_{\mathrm{med}}$ $\downarrow$ \\
\midrule
17k & 0.952M & 20.48 & 0.165 & 12.57 & 8.11 \\
23k & 1.288M & 20.64 & 0.160 & 11.78 & 8.01 \\
28k & 1.568M & 20.38 & 0.172 & 11.75 & 7.82 \\
\bottomrule
\end{tabular*}
\end{table}

\section{VLA Policy Experiment: Protocol and Full Results}
\label{app:vla}
\makeatletter
\setlength{\@fptop}{0pt}
\setlength{\@fpsep}{12pt}
\setlength{\@fpbot}{0pt plus 1fil}
\makeatother
\indent{\textbf{\textit{Setup.}}}
We compare latent-action pretraining with the released DreamDojo LAM and with \method{}. Each X-VLA policy is initialized from scratch and pretrained on the same 700{,}695 clips from EgoVerse~\citep{egoverse2026}, without robot data at this stage, to predict the latent actions extracted by its respective LAM. The 32-dimensional latent actions from each LAM are projected to 20 dimensions using the same LAM-specific PCA procedure. We keep one policy-pretraining checkpoint fixed for each LAM and repeat LIBERO post-training with three paired random seeds. Within each pair, the two policies use the same data order, optimization recipe, and 80k-update budget. Table~\ref{tab:vla-recipe} lists the complete recipe.

\indent{\textbf{\textit{Evaluation.}}}
We evaluate the official X-VLA LIBERO checkpoint and both LAM-based policies on all 10{,}030 LIBERO-Plus variants using a shared task list, initial states, environment seed, per-suite action budgets (220/280/300/520 steps), 10 settling steps, 10 denoising steps, and FP32 inference. Each policy retains its training-compatible action chunk, action interface, and proprioception. An episode succeeds only when the simulator's task predicate is met within the action budget; failed episodes remain in the denominator. Our primary LIBERO-Plus metric is overall success rate, computed as the number of successful episodes divided by 10,030. We additionally report the unweighted mean across the seven perturbation categories as a secondary metric. For the two LAM-based policies, we report the mean and sample standard deviation over three paired post-training seeds. The official X-VLA checkpoint is evaluated once on the same complete benchmark.

\indent{\textbf{\textit{Training Dynamics.}}}
Figure~\ref{fig:vla-steps} shows three paired post-training trajectories on fixed monitoring subsets. Training beyond 50k resumes from model weights with the optimizer reinitialized.
The main comparisons use the fixed 80k checkpoints evaluated on the complete benchmarks.

\indent{\textbf{\textit{Robustness Breakdown.}}}
At the fixed 80k endpoint, pretraining with \method{} outperforms pretraining with DreamDojo LAM in all three paired runs (Table~\ref{tab:vla-main}). The mean improvement is also positive in every LIBERO-Plus perturbation category, with the largest gains under object-layout, background-texture, and lighting variations (Table~\ref{tab:vla-categories}).

\begin{table}[!htbp]
\caption{\textbf{LIBERO-Plus success rate (\%) by perturbation category at 80k updates.} Mean $\pm$ sample standard deviation over three paired post-training seeds. $\Delta$ is \method{} minus DreamDojo LAM.}
\label{tab:vla-categories}
\centering\scriptsize
\setlength{\tabcolsep}{5pt}
\begin{tabular}{@{}lcccc@{}}
\toprule
Category & $n$ & DreamDojo LAM & \method{} & $\Delta$ (pp) \\
\midrule
Background textures & 1{,}076 & $67.29 \pm 0.34$ & $\mathbf{72.49 \pm 1.61}$ & $+5.20$ \\
Camera viewpoints & 1{,}599 & $92.47 \pm 0.75$ & $\mathbf{93.14 \pm 0.72}$ & $+0.67$ \\
Language instructions & 1{,}537 & $74.80 \pm 0.39$ & $\mathbf{75.71 \pm 2.00}$ & $+0.91$ \\
Lighting conditions & 1{,}142 & $88.97 \pm 0.63$ & $\mathbf{92.88 \pm 1.54}$ & $+3.91$ \\
Object layouts & 1{,}525 & $69.07 \pm 0.04$ & $\mathbf{75.39 \pm 0.30}$ & $+6.32$ \\
Robot initial states & 1{,}550 & $44.47 \pm 0.39$ & $\mathbf{45.91 \pm 0.48}$ & $+1.44$ \\
Sensor noise & 1{,}601 & $92.48 \pm 1.65$ & $\mathbf{93.28 \pm 0.81}$ & $+0.79$ \\
\bottomrule
\end{tabular}
\end{table}

\begin{table}[!htbp]
\caption{\textbf{Overall policy success and uncertainty.} Rollout-weighted success rate (\%), reported as mean $\pm$ sample standard deviation over three paired post-training seeds.}
\label{tab:vla-secondary}
\centering\scriptsize
\setlength{\tabcolsep}{7pt}
\begin{tabular}{@{}lccc@{}}
\toprule
Evaluation & DreamDojo LAM & \method{} & $\Delta$ (pp) \\
\midrule
LIBERO-Plus & $75.69 \pm 0.33$ & $\mathbf{78.25 \pm 0.37}$ & $+2.56$ \\
Standard LIBERO & $95.72 \pm 0.06$ & $\mathbf{96.30 \pm 0.51}$ & $+0.58$ \\
\bottomrule
\end{tabular}
\end{table}

\begin{table}[!htbp]

\caption{\textbf{Training recipe of the two pretrained policies} (only the LAM teacher changes, and latent coordinates and bridge are refit with the same procedure).}
\label{tab:vla-recipe}
\centering\scriptsize
\setlength{\tabcolsep}{3pt}
\renewcommand{\arraystretch}{0.92}
\begin{tabular}{@{}l>{\raggedright\arraybackslash}p{0.27\linewidth}>{\raggedright\arraybackslash}p{0.27\linewidth}>{\raggedright\arraybackslash}p{0.15\linewidth}@{}}
\toprule
Item & DreamDojo LAM teacher & \method{} teacher & Relationship \\
\midrule
Teacher & DreamDojo LAM (released 400k ckpt) & \method{} 100h & changed \\
Policy pretraining initialization & X-VLA from scratch, seed 0 &
X-VLA from scratch, seed 0 & same \\
Post-training seeds & 0, 1, 2 & 0, 1, 2 & paired \\
Downstream training length & 80{,}000 updates & 80{,}000 updates & same \\
EgoVerse pretraining clips & 700{,}695 & 700{,}695 & same rows / order / splits \\
Latent coordinates & teacher-specific 32-D \(\to\) PCA 20-D & teacher-specific 32-D \(\to\) PCA 20-D & same fitting recipe \\
Policy pretraining steps / batch & 32{,}846 / 128 & 32{,}846 / 128 & same \\
Bridge fit steps / batch / seed & 10k / 2048 / 20260720 & 10k / 2048 / 20260720 & same \\
LIBERO train / validation episodes & 1{,}592 / 120 & 1{,}592 / 120 & same \\
Downstream steps / learning rate & 80{,}000 / \(10^{-4}\) & 80{,}000 / \(10^{-4}\) & same \\
VLM / bridge LR coefficients & 0.1 / 0.1 & 0.1 / 0.1 & same \\
Trainable scope & bridge + all X-VLA & bridge + all X-VLA & same \\
Bridge auxiliary / execution weights & 1.0 / 0.3 & 1.0 / 0.3 & same \\
Camera augmentation & agent view only, \(p{=}0.7\); dropout 0.1 & agent view only, \(p{=}0.7\); dropout 0.1 & same \\
\bottomrule
\end{tabular}
\end{table}

\begin{table}[!htbp]
\centering\small
\caption{\textbf{LIBERO-Plus macro success by post-training seed at 80k updates.} Secondary evaluation using an unweighted average across the seven perturbation categories.}\label{tab:vla-seeds}
\begin{tabular}{lccc}
\toprule
LAM teacher & Seed 0 & Seed 1 & Seed 2 \\
\midrule
DreamDojo LAM & 75.89 & 75.78 & 75.29 \\
\method{} & 78.62 & 78.49 & 78.09 \\
\bottomrule
\end{tabular}
\end{table}

\paragraph{Official checkpoint evaluation and training differences.}
We evaluate the official X-VLA LIBERO checkpoint ourselves, obtaining 98.05\% on standard LIBERO and 69.74\% on LIBERO-Plus (6,995/10,030 successful episodes). The LIBERO-Plus evaluation uses the same variant list, initial states, environment seeds, action budgets, and inference-noise rules as our LAM-based policies, while retaining each policy’s native action interface. The official checkpoint uses its original pretraining and LIBERO fine-tuning recipe, with absolute end-effector actions and no camera augmentation. Our LAM-based policies use human-video latent-action pretraining, a latent-to-action bridge, and camera augmentation. Thus, the official checkpoint provides a reference under a shared evaluation protocol; the matched comparison for latent-action debiasing is CD-LAM versus DreamDojo LAM.

\subsection{RoboTwin Clean-to-Random Evaluation}
\label{app:robotwin}\label{app:vla-robotwin}
\paragraph{Evaluation setting.}
We evaluate policy robustness in the clean-to-random (C2R) setting of RoboTwin 2.0. Policies are post-trained only on demonstrations from clean environments, then evaluated in both clean and randomized scenes. Randomization changes clutter, textures, lighting, and table height, testing transfer beyond the visual conditions seen during post-training.

\paragraph{Subset construction.}
The 12-task subset was fixed from public leaderboard results before observing any results from our models. We first identify eligible tasks with official X-VLA clean-scene success of at least 80\%, randomized-scene success of at least 35\% for at least one other reported model, and an official episode horizon of at most 700 steps. Among these, we select the ten tasks with the largest X-VLA clean-to-random success drops (69--81 percentage points). We add \texttt{adjust\_bottle} and \texttt{place\_container\_plate} as anchor tasks, with official X-VLA randomized-scene success of 33--39\%, to include moderate-difficulty cases. This subset targets pronounced C2R degradation rather than average performance across all 50 RoboTwin tasks. Table~\ref{tab:robotwin_taskwise} lists all tasks and results.
\paragraph{Policy training.}
All three policies start from the same official pretrained X-VLA initialization. The latent-action policies run 20k intermediate updates using 31 robot datasets, with all RoboTwin data excluded, and EgoVerse/EgoTouch human videos occupying one third of each batch. Robot samples supervise both executable actions and latent actions; human samples supervise latent actions only. Data, sampling, optimization, and update budgets are matched between the two latent-action policies, differing only in whether the latent targets are provided by DreamDojo LAM or CD-LAM. The X-VLA baseline skips intermediate training. All policies then post-train on 50 clean demonstrations per task for 20k updates using absolute positions, 20-dimensional dual-arm actions, 30-step action chunks, three cameras, batch size 128, and learning rate $10^{-4}$. Inference directly predicts executable robot actions without latent-action inputs.

\paragraph{Rollout protocol.}
We evaluate each policy on 25 clean and 50 randomized episodes per task, using identical initial scenes across policies. Scenes are generated following the official procedure, starting from index 100000 and retaining expert-solvable instances. Task configurations are checked against the official version, and episode horizons follow the official setup. Randomized evaluation uses task names rather than the leaderboard's unseen descriptions, so leaderboard scores are not directly comparable. All policies use one post-training seed. Each latent-training policy has one simulation timeout, counted as failure. Aggregate success is the unweighted mean over the 12 tasks, equivalent to pooled episode success within each scene condition because evaluation counts are equal across tasks.

\paragraph{Statistical analysis.}
We compute paired 95\% confidence intervals using 2000 bootstrap replicates, resampling matched episodes within each task. CD-LAM improves randomized-scene success over DreamDojo LAM by 5.0 percentage points, with interval $[1.8,8.2]$; discordant successes are 62 versus 32 episodes. Its clean-scene difference is 2.7 points, with interval $[-0.7,6.0]$. Relative to skipping intermediate training, the randomized gain is 8.8 points, with interval $[5.3,12.3]$. These confidence intervals quantify rollout variability for the evaluated tasks and fixed trained policies.

\begin{table}[!htbp]
\centering\small
\caption{\textbf{Task-level success rates (\%) on RoboTwin C2R.} All matched policies use the same initial scenes: 25 clean and 50 randomized episodes per task. X-VLA skips intermediate training; LAM and CD-LAM use matched intermediate training.}
\label{tab:robotwin_taskwise}
\setlength{\tabcolsep}{5pt}
\begin{tabular}{@{}lrrr|rrr@{}}
\toprule
& \multicolumn{3}{c|}{Clean} & \multicolumn{3}{c}{Randomized} \\
Task & X-VLA & LAM & CD-LAM & X-VLA & LAM & CD-LAM \\
\midrule
lift\_pot & 96 & 100 & 100 & 30 & 24 & 38 \\
grab\_roller & 100 & 100 & 100 & 52 & 42 & 60 \\
rotate\_qrcode & 84 & 88 & 84 & 22 & 18 & 26 \\
place\_phone\_stand & 64 & 76 & 84 & 4 & 4 & 4 \\
move\_playingcard\_away & 92 & 80 & 76 & 18 & 18 & 12 \\
beat\_block\_hammer & 88 & 100 & 96 & 8 & 10 & 18 \\
place\_object\_basket & 72 & 36 & 52 & 2 & 16 & 20 \\
place\_empty\_cup & 84 & 88 & 88 & 32 & 40 & 48 \\
place\_bread\_skillet & 72 & 68 & 76 & 6 & 6 & 4 \\
open\_laptop & 80 & 92 & 84 & 38 & 62 & 58 \\
adjust\_bottle & 100 & 100 & 100 & 62 & 62 & 66 \\
place\_container\_plate & 96 & 80 & 100 & 26 & 44 & 52 \\
\midrule
Mean & 85.7 & 84.0 & 86.7 & 25.0 & 28.8 & 33.8 \\
\bottomrule
\end{tabular}
\end{table}

\end{document}